\documentclass[11pt]{article}

\usepackage[margin=1in]{geometry}
\usepackage{amsmath,amssymb}
\usepackage{graphicx}
\usepackage{xcolor}
\usepackage{hyperref}
\usepackage{microtype}   
\usepackage{tikz}
\usetikzlibrary{arrows.meta,positioning,shadows.blur,calc,shapes.geometric}
\usepackage{tcolorbox}
\usepackage{subcaption}
\usepackage{amsthm}

\usepackage{titlesec}
\titlespacing*{\section}{0pt}{*1.5}{*1}
\titlespacing*{\subsection}{0pt}{*1.2}{*0.8}
\titlespacing*{\paragraph}{0pt}{*0.8}{1em}

\usepackage{enumitem}
\setlist{nosep} 

\AtBeginDocument{
  \setlength{\abovedisplayskip}{3pt plus 1pt minus 1pt}
  \setlength{\belowdisplayskip}{3pt plus 1pt minus 1pt}
  \setlength{\abovedisplayshortskip}{0pt plus 1pt}
  \setlength{\belowdisplayshortskip}{2pt plus 1pt minus 1pt}
}

\newtheorem{hypothesis}{Hypothesis}
\newtheorem{prediction}{Prediction}
\newtheorem{definition}{Definition}
\newtheorem{lemma}{Lemma}
\newtheorem{proposition}{Proposition}

\definecolor{labDark}{RGB}{20,20,22}
\definecolor{labBlue}{HTML}{9f0796}    
\definecolor{labPurple}{HTML}{9f0796} 
\definecolor{labCyan}{HTML}{cb0ac0}    
\definecolor{labAccent}{HTML}{6e0066}  
\definecolor{labGray}{RGB}{142,142,147}
\definecolor{labLight}{RGB}{249,249,252}
\definecolor{labRed}{RGB}{255,59,48}

\tikzset{
  labAxis/.style={draw=labDark!50, line width=0.7pt, -{Stealth[round, length=2.8mm, width=1.8mm]}},
  labGrid/.style={draw=labDark!5, line width=0.3pt},
  labData/.style={line width=2.4pt, draw=labBlue, line cap=round, line join=round},
  labLegacy/.style={line width=1.2pt, draw=labDark!25, dashed},
  labFill/.style={fill=labBlue!10, opacity=0.4},
  labNode/.style={
    draw=labDark!12, 
    line width=0.1pt,
    fill=white, 
    rounded corners=8pt, 
    inner xsep=14pt, 
    inner ysep=12pt, 
    blur shadow={shadow blur steps=8, shadow opacity=12, shadow xshift=0.2pt, shadow yshift=-0.8pt},
    font=\sffamily\small
  },
  labArrow/.style={-{Stealth[round, length=2.2mm, width=1.6mm]}, line width=1pt, draw=labBlue!80}
}

\title{The Headless Firm: How AI Reshapes Enterprise Boundaries}
\author{
  Tassilo Klein \\
  \textit{Mantix} \\
  \texttt{tk@mantix.cloud}
  \and
  Sebastian Wieczorek \\
  \textit{Mantix} \\
  \texttt{sw@mantix.cloud}
}
\date{}

\begin{document}
\maketitle

\let\thefootnote\relax\footnotetext{\textbf{Conflict of Interest:} The authors are founders of Mantix (mantix.cloud), a company building infrastructure in the agentic enterprise space described in this paper. This paper represents independent academic work; no commercial claims are made herein.}

\begin{abstract}
The boundary of the firm is determined by coordination cost. We argue that agentic AI induces a structural change in how coordination costs scale: in prior modular systems, integration cost grew with interaction topology ($O(n^2)$ in the number of components); in protocol-mediated agentic systems, integration cost collapses to $O(n)$ while verification scales with task throughput rather than interaction count. This shift selects for a specific organizational equilibrium---the Headless Firm---structured as an hourglass: a personalized generative interface at the top, a standardized protocol waist in the middle, and a competitive market of micro-specialized execution agents at the bottom. We formalize this claim as a coordination cost model with two falsifiable empirical predictions: (1) the marginal cost of adding an execution provider should be approximately constant in a mature hourglass ecosystem; (2) the ratio of total coordination cost to task throughput should remain stable as ecosystem size grows. We derive conditions for hourglass stability versus re-centralization and analyze implications for firm size distributions, labor markets, and software economics. The analysis predicts a domain-conditional Great Unbundling: in high knowledge-velocity domains, firm size distributions shift mass from large integrated incumbents toward micro-specialized agents and thin protocol orchestrators.
\end{abstract}


\section{Introduction}
The boundary of the firm is set by the cost of coordination. As Coase argued, firms expand when internal direction is cheaper than market exchange, and contract when market transaction costs fall. \cite{coase1937nature} The critical variable is how coordination costs scale: whether they grow primarily with the number of interaction edges between components, or primarily with task throughput.
This paper advances a specific, testable claim: agentic AI can change the scaling regime of coordination. In prior modular systems (SOA, microservices), coordination cost grew with interaction topology—interfaces, release coupling, governance dependencies—often approaching a combinatorial surface. In agentic systems, by contrast, a larger share of integration work shifts from bilateral negotiation toward protocol-mediated tool access and outcome-based verification. If that shift holds, coordination can scale primarily with task throughput rather than with the number of pairwise integration edges.
When coordination scales with throughput, modular delegation becomes economically sustainable. Execution can fragment into micro-specialized agents without triggering the superlinear coordination overhead that historically made modular systems economically unviable at scale. At the same time, fragmentation at the execution layer creates a usability problem: humans cannot navigate an exploding interface surface area. The architectural response is an hourglass equilibrium: a personalized intent layer (Generative UI, or GenUI) at the top, a thin protocol-and-policy waist that compiles intent into constrained actions, and a competitive market of vertical execution agents at the bottom.
\begin{definition}[Coordination Scaling Regime]
We define a coordination regime as topology-dominated when coordination cost scales with the number of interaction edges between components, and throughput-dominated when cost scales primarily with the number of tasks executed. The Headless Firm emerges when agentic systems shift coordination from the former regime to the latter.
\end{definition}
We call the organizational analogue of this equilibrium the \emph{Protocol-Coordinated Firm (Headless Firm)}: a structure in which execution is decoupled from centralized management and coordinated via standardized protocols and verification rather than hierarchical supervision. The headless firm is ``headless'' in the specific sense that the user-facing intent surface and the execution surface are separated by a thin coordination layer; it is not a claim that governance disappears, but that it is compressed into standard contracts, policy gates, and evaluation.
Our contributions are as follows. (1) We propose a coordination cost model that formalizes the distinction between topology-dominated and throughput-dominated coordination regimes (Section \ref{sec:model}). (2) We derive two falsifiable empirical predictions from this model. (3) We characterize stability conditions for the hourglass equilibrium and identify four distinct re-centralization pathways (Section \ref{sec:discussion}). (4) We derive labor market and firm size implications conditional on knowledge velocity (Section \ref{sec:implications}). The central claim is conditional: the hourglass is the stable organizational form if and only if per-task verification cost $v(k)$ grows sublinearly in workflow width $k$.
This decomposition is not frictionless. The history of service-oriented architectures demonstrates that modular systems fail when coordination costs scale superlinearly with interaction edges. Latency, partial failures, schema drift, identity and compliance layers, and system-of-record gravity all reintroduce integration burdens. The burden of proof therefore lies in whether agentic modularity meaningfully changes the coordination scaling law: does verification and governance remain a throughput problem, or does it re-expand into a topology problem (blame allocation, cascading failures, bespoke policy)?
If the new scaling regime holds, value concentrates at the architectural edges. Protocol coordination commoditizes into a thin waist, while differentiation shifts upward into personalized intent and downward into deep vertical execution. If it fails—because protocols thicken into proprietary gatekeepers, verification becomes topology-dependent, or liability centralizes—re-centralization follows. The argument of this paper is therefore conditional but concrete: \emph{a change in the scaling law of coordination selects for an hourglass architecture, and that architecture shifts enterprise boundaries toward a headless firm.}

\section{Related Work}
\paragraph{Theory of the firm and transaction costs.} Our work builds on Coase's \cite{coase1937nature} transaction cost theory of the firm and Williamson's \cite{williamson1981economics} asset specificity extension. Prior work has examined how digital technologies affect firm boundaries: Brynjolfsson and Hitt \cite{brynjolfsson2000beyond} document IT-driven decentralization; Garicano and Rossi-Hansberg \cite{garicano2006organization} model knowledge hierarchies and show that communication cost reductions lead to flatter organizations. We extend this line by introducing the knowledge decay rate as a second dimension that interacts with coordination cost to determine optimal firm structure.
\paragraph{Modular architectures and platform economics.} Baldwin and Clark \cite{baldwin2000design} establish the modularity thesis; Langlois \cite{langlois2002modularity} extends it to dynamic capabilities. Our hourglass equilibrium is related to the layered protocol model of Akhshabi and Dovrolis \cite{akhshabi2011evolution} and to the end-to-end principle \cite{saltzer1984end}. The ``thick waist'' risk we identify in Section \ref{sec:thickwaist} is related to platform envelopment \cite{eisenmann2006strategies} and analysis of multi-sided markets \cite{rochet2003platform}.
\paragraph{Multi-agent AI systems.} Recent work on tool-augmented language models \cite{schick2024toolformer} and agent benchmarks \cite{qin2023toolllm, liu2024agentbench} establishes empirical baselines for agent capability. The MCP protocol \cite{anthropic2024mcp} provides the technical substrate our protocol waist assumes. Concurrent work on agentic delegation \cite{tomasev2026intelligent} and AI operating systems \cite{mei2025aios} addresses related coordination problems at the systems level. Our contribution is to bring these systems-level developments into contact with organizational economics.
\paragraph{Labor and automation.} Autor et al. \cite{autor2020superstar} on superstar firms and Acemoglu and Restrepo \cite{acemoglu2022tasks} on automation and labor demand provide the empirical backdrop for our Section \ref{sec:implications} analysis. Song et al. \cite{song2019firming} on between-firm wage dispersion is directly relevant to our scalable boutique prediction.
\section{The Economic and Temporal Imperative}
The hourglass thesis depends on two independent but interacting forces. First, AI reduces the spatial cost of coordinating across firm boundaries, altering the traditional transaction-cost calculus. Second, accelerating knowledge decay increases the temporal cost of maintaining broad internal capabilities. Together, these forces create the economic pressure that selects for modular, protocol-mediated firms.
\subsection{Transaction-cost theory}
Coase's theory of the firm explains vertical integration as a response to the costs of using markets (discovering prices, negotiating contracts, and enforcing agreements) \cite{coase1937nature}. If AI agents reduce these costs, then market-based coordination becomes more attractive, and the efficient firm boundary may shrink. Importantly, this reduction is unlikely to be uniform: stateful dependencies, governance requirements, and SoR switching costs can keep large parts of the stack effectively ``integrated'' even when execution is modularized. This aligns with Williamson's analysis of asset specificity, where highly specific investments (like complex SoR state) encourage integration to prevent hold-up problems \cite{williamson1981economics}.

\subsection{The temporal constraint: knowledge decay}
While Coase explains why firms shrink when spatial coordination costs fall, he does not account for the temporal stability of the firm's internal processes. Historically, firms integrated vertical capabilities because best practices were stable; standard operating procedures could be codified and amortized over years.
Today, high-stakes domains (medicine, law, compliance, software engineering) face an accelerating burden of knowledge. As Jones (2009) demonstrates, the expanding knowledge frontier forces individuals to narrow their expertise to reach the cutting edge \cite{jones2009burden}. This necessitates a shift from individual generalists to specialized teams---or, in our framework, specialized agents. This induces a Red Queen effect \cite{vanvalen1973redqueen}: the firm must run ever faster simply to maintain its current relative standing. In this high-velocity regime, the integrated firm faces a ``complexity wall'' \cite{tainter1988collapse}: the cost of constantly rewriting internal tooling to match the frontier of knowledge exceeds the cost of purchasing execution from specialized external agents who amortize that maintenance across many clients. On this view, headless structure is attractive when two gradients steepen at once: market coordination becomes cheaper to invoke, while maintaining broad internal capabilities becomes more expensive as domains churn.
However, the same logic cuts both ways: AI can also reduce \emph{internal} coordination costs---automating monitoring, reporting, and information routing---which in isolation would support \emph{larger} firms. The key question is therefore not whether AI reduces coordination, but \emph{which kind}: transaction costs across firm boundaries, or complexity costs within them.
On the internal side, even with AI-assisted management, organizations face diseconomies of scale rooted in complexity: innovation slows, adaptation lags, and incentive alignment degrades as firm size grows \cite{west2017scale}. This mirrors Simon's ``bounded rationality''---the observation that no central planner, human or AI, can process all the distributed information needed to coordinate a large, fast-moving system \cite{simon1955behavioral}. The principal--agent problem \cite{jensen1976theory} persists: ensuring that delegated agents act in the principal's interest still requires monitoring, and those monitoring costs can offset the benefits of delegation.
More fundamentally, AI does not eliminate what Hayek called the ``knowledge calculation problem'' \cite{hayek1945knowledge}: decentralized systems adapt to local signals faster than any optimized monolith. As domain knowledge accelerates---medical guidelines, legal precedent, software tooling---the cost of keeping internal processes current rises superlinearly \cite{wuchty2007increasing}, while external specialists can amortize that cost across many clients.

\subsection{Empirical and theoretical constraints on firm growth}
Empirically, firm sizes follow heavy-tailed distributions; for the United States, Axtell reports a Zipf-like distribution over firm sizes \cite{axtell2001zipf}.

\begin{hypothesis}[Domain-Conditional Unbundling]
In sectors where (i) knowledge velocity is high, (ii) system-of-record switching costs are low, and (iii) protocol standardization is achievable, lower coordination friction shifts the firm size distribution toward smaller, specialized producers. This prediction does not apply uniformly: sectors with stable knowledge functions, high capital intensity, or strong network effects may experience consolidation rather than fragmentation. As AI agents reduce the transaction costs of external coordination, probability mass shifts from the ``head'' (large integrated firms) to the ``tail'' (micro-specialized entities). In a simple power-law view of firm sizes, this corresponds to a steepening of the exponent \cite{gabaix2009power}, such that the aggregate output of the long tail begins to rival that of the monolithic head.
\end{hypothesis}

\begin{figure}[ht]
  \centering
  \begin{tikzpicture}[font=\sffamily\footnotesize]
    \foreach \x in {0.5,1.0,...,9.5} 
      \foreach \y in {0.5,1.0,...,4.5}
        \fill[labDark!15] (\x,\y) circle (0.4pt);

    \draw[labAxis] (0,0) -- (10.2,0) node[right, labDark!40] {firm size (log)};
    \draw[labAxis] (0,0) -- (0,4.8) node[above, labDark!40] {frequency (log)};

    \draw[labLegacy, tension=0.8] plot[smooth] coordinates {(0.8,4.15) (3,3.2) (6,2.2) (9.3,1.3)};
    \node[labDark!50, font=\sffamily\bfseries\tiny, anchor=west] at (9.4, 1.3) {Monolithic Era};

    \draw[labData, tension=0.85] plot[smooth] coordinates {(0.8,4.65) (2.2,4.05) (3.5,2.65) (5.2,1.55) (7.2,1.15) (9.3,0.75)};
    \node[labBlue, font=\sffamily\bfseries\tiny, anchor=west] at (9.4, 0.75) {Agentic Era};

    \fill[labBlue!15, opacity=0.3] (0.8,4.15) -- (0.8,4.65) -- (2.2,4.05) -- (3,3.2) -- cycle;
    \draw[labArrow, draw=labBlue, -{Stealth[round, scale=0.9]}, line width=1.2pt] (1.5, 3.5) -- (1.5, 4.35);
    \node[labBlue, font=\sffamily\bfseries\tiny, anchor=north, align=center] at (1.5, 3.4) {Unbundling:\\more verticals};

    \fill[labPurple!15, opacity=0.3] (6.0,2.2) -- (9.3,1.3) -- (9.3,0.75) -- (7.2,1.15) -- cycle;
    \draw[labArrow, draw=labPurple, -{Stealth[round, scale=0.9]}, line width=1.2pt] (8.0, 1.8) -- (8.0, 1.0);
    \node[labPurple, font=\sffamily\bfseries\tiny, anchor=south, align=center] at (8.0, 1.9) {Efficiency:\\fewer employees};

    \fill[labDark!20, opacity=0.45]
      (3.2,3.05) -- (6.0,2.2) -- (7.6,1.75) -- (7.6,1.15) -- (5.2,1.55) -- (3.2,2.85) -- cycle;
    \node[labDark!80, font=\sffamily\bfseries\tiny, anchor=north] at (5.9, 2.0) {Missing middle};

  \end{tikzpicture}
  \caption{The hypothesized deformation of firm sizes (The Great Unbundling). Two distinct effects reshape the distribution: (1) {\bf Unbundling}---falling transaction costs enable proliferation of micro-specialized verticals; (2) {\bf Efficiency}---AI-enabled automation allows large firms to operate with fewer employees and reduced service requirements. Both mechanisms contribute to a flattened distribution, consistent with West's analysis of hierarchical scaling constraints in organizations \cite{west2017scale}. We provide the formal economic derivation of this shift—demonstrating why knowledge decay overrides AI's span-of-control benefits—in Section 9.1.}
\end{figure}
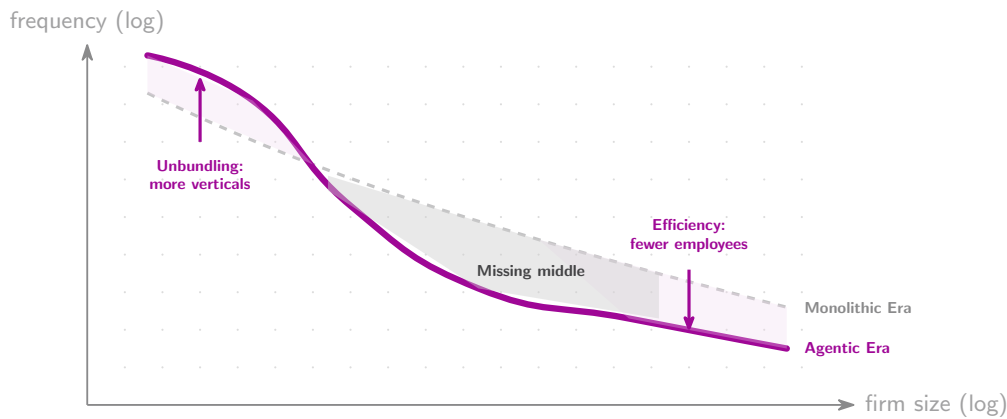

The term ``unbundling'' has been used in enterprise software contexts to describe the shift from monolithic suites to specialized vertical solutions, a digital manifestation of Stigler's lifecycle of vertical disintegration \cite{stigler1951division, casado2024unbundling}. We adopt this term but extend it to the \emph{boundary of the firm itself}: we argue that unbundling is driven not only by the proliferation of micro-specialists (the standard narrative), but also by efficiency-driven downsizing of large incumbents---two distinct forces that together reshape the firm size distribution.
Several countervailing forces may constrain this unbundling in practice.
\paragraph{Platform concentration.} AI infrastructure providers (foundation model vendors, cloud hyperscalers) may exhibit winner-take-most dynamics due to scale economies in training and network effects in ecosystems, potentially creating new large firms even as application-layer firms fragment.
\paragraph{Regulatory and capital requirements.} Industries with high compliance burdens (finance, healthcare) or significant capital intensity (manufacturing, infrastructure) impose minimum viable firm sizes that resist full disaggregation.
\paragraph{Network effects and marketplace dynamics.} Certain verticals may consolidate around dominant platforms that aggregate demand or supply, leading to bimodal rather than uniformly flattened distributions. These boundary conditions suggest that the Great Unbundling will be uneven across sectors, with the most pronounced effects in knowledge work, low-capital services, and domains where SoR lock-in is weakest.
This long tail is also consistent with the rise of ``gig economy'' and marketplace coordination as a pre-GenAI precursor. As Sundararajan \cite{sundararajan2016sharing} observed, these platforms lowered matching costs and blurred traditional boundaries, yet they did not entirely dissolve the firm. A plausible reason is that many transaction costs remained stubbornly human (specification, monitoring, dispute resolution). If agentic systems reduce these frictions, they can unlock a more general version of marketplace coordination---the ``Coasean singularity'' argument \cite{shahidi2025coasean}.
Theoretically, Lucas models limits to firm growth via diminishing returns to managerial talent and increasing coordination costs \cite{lucas1978size}. These results motivate a ``complexity wall'': beyond some scale, additional headcount adds disproportionate overhead rather than productive capacity.

\section{Lessons from SOA and the Limits of Prior Modularity}
Enterprise software has already lived through a large-scale attempt at modularization via service-oriented architectures (SOA) and microservices. That experience is a useful prior: exposing capabilities as services made ownership boundaries clearer, but it also surfaced new coordination costs that often offset the benefits, a phenomenon often described as the ``microservices premium'' \cite{fowler2014microservices}.

\subsection{Practical frictions in SOA}
SOA deployments repeatedly surfaced the same family of integration burdens. Distributed systems costs arise because network latency, partial failures, and cascading outages shift failure modes from deterministic to probabilistic \cite{deutsch1994fallacies}. Interface drift compounds this: schemas and API contracts evolve continuously, turning backward compatibility into a permanent maintenance tax. More insidiously, services can agree on a field name while disagreeing on its meaning---a ``semantic mismatch'' that reveals the limits of pure syntactic coordination \cite{evans2004domain}. Finally, cross-service invariants require distributed transactions or compensating workflows, and identity, logging, and tracing tend to re-centralize into a platform layer, recreating coordination overhead under a new name.
A particularly instructive failure is the ``Smart Pipe'' trap. Previous orchestration efforts---most notably the Enterprise Service Bus (ESB)---collapsed because they attempted to resolve coordination \emph{syntactically}: rigid canonical schemas and fragile XSLT transformations broke whenever any endpoint changed \cite{hohpe2003enterprise}. Agentic orchestration avoids this failure mode because it coordinates \emph{semantically}. Large language models act as fuzzy universal adapters that negotiate intent and map data across disparate schemas without brittle, hand-coded translators \cite{chappell2004enterprise}. The hourglass architecture converts these historical lessons into design rules: the orchestrator owns versioned tool contracts and compatibility shims; shared evaluation harnesses push semantic alignment to the protocol layer; writes against the system of record are mediated through reversible, policy-gated workflows. Governance and observability remain concentrated in the thin waist rather than scattered across agents.

\subsection{What agentic orchestration can overcome}
Agentic systems do not repeal distributed-systems constraints, but they can substantially reduce the \emph{human} costs of managing them. Three mechanisms are central.

\textbf{Lower specification cost.} Intent-first interaction turns under-specified requests into executable plans through iterative clarification, realizing Postel's Robustness Principle \cite{postel1981tcp}: agents accept ambiguous natural-language inputs while emitting conservative, structured tool calls.
\textbf{Generated integration glue and model-driven adapter maintenance.} Agents can synthesize connectors, schema mappings, and migration scripts, and update them autonomously as APIs drift. Benchmarks such as ToolBench \cite{qin2023toolllm} confirm that tool selection and schema mapping can be learned and evaluated at scale, reframing integration as an engineering problem rather than an open-ended negotiation.
\textbf{Continuous evaluation and routing.} Orchestrators can detect breakage, propose patches, and comparatively score services using outcome-based tests---supporting a market of interchangeable executors without manual re-integration.

\subsection{Sublinear Coordination Hypothesis}

The central claim of this paper can be stated as a testable economic hypothesis:
\begin{hypothesis}[Sublinear Coordination Efforts]
In systems where execution is delegated to AI-mediated agents, total coordination cost grows more slowly than system complexity, provided that:
(i) tool interaction is standardized at a shared protocol boundary, and
(ii) verification and integration are shifted from bilateral negotiation to model-driven inference evaluated through scalable, outcome-based tests.
\end{hypothesis}
If integration edges are collapsed at the protocol layer ($E = \mathcal{O}(n)$) and per-task verification cost satisfies $v(k) = \mathcal{O}(k^\gamma)$ for $\gamma < 1$, then total coordination cost grows sublinearly with system complexity for fixed task distribution. This condition fails---and hourglass architecture loses its advantage---when cross-provider coupling intensity $q_t$ grows with ecosystem size, forcing the quadratic term in equation (\ref{eq:verification_coupling}) to dominate. Empirically, $\gamma < 1$ requires that evaluation harnesses, policy templates, and outcome tests are reusable across agent compositions rather than constructed independently per task.

\section{A Coordination Cost Model of the Hourglass Equilibrium}
\label{sec:model}
The preceding sections argued qualitatively that agentic systems can alter the scaling law of coordination. We now formalize that claim in a minimal economic model. The purpose of this model is not to capture all organizational complexity, but to isolate the structural condition under which the hourglass architecture becomes economically stable.
\subsection{Coordination scaling: a minimal model}
We make the scaling claim explicit with a minimal cost model.

\paragraph{Objects.}
Let $n$ denote the number of independent execution providers (agents/services/components). Over a period, let $T$ be the number of tasks executed. Each task $t$ touches a subset of providers; let $k_t$ be the number of providers involved in task $t$ (the \emph{workflow width}). Let $E$ denote the number of maintained integration edges---adapters, tool contracts, schema mappings, and compatibility shims---required for the system to function.

\paragraph{Coordination cost decomposition.}
Define total coordination cost over the period as
\begin{equation}
C \;=\; C_{\text{integration}}(E) \;+\; C_{\text{verification}}(T,\{k_t\}) \;+\; C_{\text{governance}}(T),
\label{eq:coord_cost_decomp}
\end{equation}
where $C_{\text{integration}}$ is the cost of maintaining and versioning integration edges (API contracts, schema mappings, compatibility shims); $C_{\text{verification}}$ is the cost of checking that task outputs satisfy correctness and policy requirements across the $k_t$ involved providers; and $C_{\text{governance}}$, which we treat as linear in throughput by assumption (audit and routing-policy checks per task, not per integration pair), scales identically in both regimes and therefore does not determine which is preferable---the key question is whether $C_{\text{integration}}$ and $C_{\text{verification}}$ escape superlinear growth.

\paragraph{Two coordination regimes.}
In topology-dominated modularity (a common SOA/microservices failure mode), integration cost is edge-driven: $C_{\text{integration}}(E) = aE$, where, with weak protocol discipline, $E = \Theta(n^2)$ due to the combinatorial pairwise-edge growth famously formalized by Brooks~\cite{brooks1975mythical}, implying a superlinear integration term.
In a protocol-mediated hourglass, pairwise integration is collapsed into a shared waist so each provider integrates once to the protocol boundary: $E = \Theta(n)$, provided protocol evolution remains backward-compatible and compatibility shims are centralized at the waist. Consequently, $C_{\text{integration}}$ scales approximately linearly in ecosystem size.
\paragraph{Verification as the decisive term.}
With a protocol waist, the dominant question becomes how verification scales with workflow width. We write per-task verification as
We separate per-task verification into two components:
\begin{equation}
v(k) = v_{\text{local}}(k) + v_{\text{coupling}}(k)
\label{eq:verification_sum}
\end{equation}
where $v_{\text{local}}$ represents the cost of checking constraints against reusable harnesses, and $v_{\text{coupling}}$ represents the cost of checking cross-provider invariants. The hourglass architecture requires that $v_{\text{local}}(k)$ grows slowly in workflow width $k$. We formalize this relying on the following condition:

\begin{lemma}[Verification Reuse Condition]
\label{eq:lemma1}
Suppose tasks are evaluated using a library of reusable outcome tests and policy templates drawn from a fixed domain vocabulary. If the marginal cost of extending this library grows sublinearly with workflow width, then the local per-task verification cost satisfies $v_{\text{local}}(k) = \mathcal{O}(k^\gamma)$ for some $\gamma < 1$.
\end{lemma}
This condition emerges when evaluation artifacts are shared across workflows, so extending the library amortizes across many compositions rather than scaling with each additional provider. The exponent $\gamma$ can be interpreted as the inverse reuse factor of verification artifacts: when $\gamma \approx 0$, a single reusable test harness suffices; when $\gamma \to 1$, each additional provider requires proportionally more verification work. The condition $\gamma < 1$ is an essential driver of the hourglass architecture's efficiency. Intuitively, the hourglass architecture is efficient when most correctness checks can be applied independently to individual agents using reusable evaluation artifacts, and only a minority of tasks require enforcing tight consistency across multiple providers.
Verification becomes topology-like again when correctness depends on cross-provider invariants ($v_{\text{coupling}}$). Let $q_t \in [0,1]$ denote \emph{coupling intensity} for task $t$ (e.g., the probability that task $t$ triggers cross-provider reconciliation). A minimal way to express total verification cost over $T$ tasks is:
\begin{equation}
C_{\text{verification}} \;=\; \sum_{t=1}^{T} \left( b v_{\text{local}}(k_t) \;+\; d\, v_{\text{coupling}}(k_t) \right) \;=\; \sum_{t=1}^{T} \left( b k_t^\gamma \;+\; d\, q_t\, k_t^2 \right),
\label{eq:verification_coupling}
\end{equation}
The quadratic form of the coupling term follows from the assumption that cross-provider invariant checking requires pairwise consistency verification between each pair of coupled providers involved in a task. If task $t$ triggers coupling between $m_t \le k_t$ providers, and each pair requires independent reconciliation, verification work scales as $\mathcal{O}(m_t^2)$. We simplify by letting $q_t = m_t/k_t$ denote the fraction of involved providers that are coupled, giving a coupling term proportional to $q_t \cdot k_t^2$. The quadratic form represents a worst-case bound when invariants require pairwise reconciliation; hierarchical or sampled verification schemes reduce the exponent but do not eliminate topology dependence.

\begin{proposition}[Hourglass Instability]
If coupling intensity grows proportionally with ecosystem size, i.e., $q_t = \Theta(n)$, then verification cost becomes quadratic and the hourglass architecture loses its coordination advantage.
\end{proposition}

\paragraph{Scope condition.}
This instability proposition sharply bounds the scope of the theory. We predict hourglass stability in read-heavy or reversible-write workflows where coupling intensity remains low. Conversely, the hourglass model fails in domains requiring tightly coupled state machines, safety-critical control systems, or high-invariant core ledgers, where coupling forces topology-dependent verification.

\subsection{Observability as a Precondition}
The coordination cost model in Section 5.1 treats workflows as machine-addressable objects: it assumes that n, T, and $k_t$ are measurable, that integration edges E can be maintained, and that verification harnesses can be applied to task outputs. These assumptions are not automatically satisfied. In practice, the majority of enterprise processes exist only implicitly — encoded in UI interaction sequences, human memory, and undocumented branching logic that no API exposes. Process observability is therefore not a downstream engineering concern but a logical precondition of the model itself: without a machine-readable representation of how workflows actually operate, the coordination cost decomposition in equation (1) cannot be instantiated.
This gap is structural rather than incidental. Enterprise applications were designed to expose data models, not process topology. The reachable state space as traversed in practice — including exception paths, permission-conditioned branches, and context-dependent defaults — is systematically absent from any API-level representation.
For the hourglass model to be instantiated, any process observability primitive must satisfy three abstract conditions: (i) topological completeness — the reachable state space as actually traversed must be represented, not only the nominal path; (ii) semantic interpretability — transitions must carry intent, not merely syntactic descriptors, so the orchestration layer can reason about what a step means rather than only what it does; and (iii) distributional awareness — traversal frequencies must be captured, so that normal-flow paths are distinguishable from exception cases. This last condition is necessary for the verification reuse condition in Lemma \ref{eq:lemma1}: reusable harnesses presuppose sufficient knowledge of workflow structure to identify which compositions are common and which are exceptional. These conditions cannot be satisfied at the API layer alone; the delta between what an application exposes programmatically and how users actually traverse it is systematically absent from any server-side representation.
The gap these conditions expose defines a necessary infrastructure layer that is not yet standardized — analogous to the role distributed tracing played as a precondition for microservices observability. Until such a layer exists, the coordination cost decomposition in equation (1) remains a theoretical construct rather than an operationalizable framework.

\subsection{Falsifiable predictions}
\label{sec:falsifiable}
The model yields two empirical predictions that can be evaluated with production telemetry and engineering effort accounting.

\begin{prediction}[Edge-addition marginal cost]
Let $\Delta C_{\text{integration}}$ be the incremental integration work required to add one new execution provider. Without a protocol waist, adding a provider induces new pairwise edges ($\Delta E = \Theta(n)$), so $\Delta C_{\text{integration}} \propto n$. With a protocol waist, the new provider integrates once to the protocol boundary ($\Delta E = \Theta(1)$), so $\Delta C_{\text{integration}} \approx \text{constant}$.
\end{prediction}

\begin{prediction}[Verification stability under ecosystem growth]
In a stable hourglass, $C/T$ should remain approximately stable or grow slowly with $n$---not proportionally to $n$ or $n^2$. Conversely, if coupling intensity $q$ rises with ecosystem size (e.g., due to stateful cross-provider invariants), then \eqref{eq:verification_coupling} predicts that $C/T$ will increase sharply with $n$ via the $k_t^2$ term, signaling hourglass collapse and re-centralization pressure.
\end{prediction}
In the agentic setting, $n$ is the number of vertical agents, $k_t$ is the workflow width of a single user task, $E$ represents maintained tool contracts and schema mappings, and $q_t$ reflects the density of cross-agent invariants and write-authority constraints. The model provides a selection condition: the hourglass is stable when $E = \Theta(n)$ and $q_t$ remains low; it collapses toward re-centralization when coupling forces the quadratic term in \eqref{eq:verification_coupling} to dominate. This selection condition is the economic logic behind Sections~\ref{sec:execution} and~\ref{sec:interaction}: vertical specialization achieves low $\gamma$ by staying within a narrow domain; generative intent keeps $k_t$ manageable by compiling user goals rather than exposing raw agent interfaces.
\subsection{What agents do not overcome (and may worsen)}
Several frictions remain binding:
\begin{itemize}
  \item \textbf{System-of-record lock-in:} Switching SoR is costly due to migration, reconciliation, downstream dependencies, and audit risk. This mirrors Moore's distinction between systems of record (which resist change) and systems of engagement (which require fluidity) \cite{moore2011systems}.

    However, the monolithic nature of the SoR is increasingly contested by decentralized data architectures:
    \begin{itemize}
        \item \textbf{Composable ERP:} Frameworks emphasizing the decoupling of core business logic into modular, best-of-breed microservices rather than relying on vendor-locked suites \cite{gartner2020composable}.
        \item \textbf{Zero-copy Federation:} Open table formats (e.g., Apache Iceberg) and data lakehouses \cite{apache2024iceberg} attempt to decouple canonical data state from proprietary application logic.
    \end{itemize}
    Historically, these unbundling initiatives have stalled due to severe integration complexity, network latency during distributed joins, and failures in scaling manual data governance \cite{gartner2025erp}. In the context of the headless firm, agentic orchestration offers a novel mechanism to overcome these historic frictions: agents can automate semantic alignment across fragmented microservices and autonomously monitor data quality across distributed environments, potentially enabling the modularization of the SoR itself.
  \item \textbf{Write authority and liability:} delegating writes requires robust authorization, approvals, provenance, and replayability. This creates a classic principal-agent problem where the agent's objective function may not perfectly align with the principal's intent, a challenge framed in economics as incomplete contracting \cite{hart1990property, hadfield2019incomplete}.
  \item \textbf{Tail risks and debugging:} stochastic behavior shifts failure from ``wrong response code'' to ``wrong outcome,'' raising the burden of monitoring and dispute resolution. These failures often manifest as negative side effects or reward hacking \cite{amodei2016concrete, zhou2024webarena, liu2024agentbench, deng2023mind2web}, requiring scalable oversight mechanisms beyond simple unit tests.
\end{itemize}
To manage these stochastic tail risks programmatically, the orchestration layer requires robust evaluation harnesses. Current benchmarks rely on fixed app environments, obscuring how reliability degrades under real-world variation: empirical work shows task success fluctuating by more than 50\% across app design and content variations, with failure modes such as looping and hallucinating actions varying strongly by environment \cite{openapps2026}.

\subsection{Three illustrative configurations}
The stability condition derived above depends critically on the intensity of cross-provider coupling and the rigidity of the underlying system of record. To make this dependency concrete, we outline three stylized configurations that vary in state density and invariant strictness. These cases illustrate where the hourglass is robust, conditionally stable, or structurally constrained.
\paragraph{Hourglass with a dominant SoR (swappable execution).} A CRM or ERP remains the canonical record, but many specialized agents (e.g., enrichment, outreach drafting, ticket triage) can be swapped because they mainly \emph{read} from the SoR and propose changes that are mediated by the orchestrator via human-in-the-loop or policy-gated writes.
\paragraph{Hourglass in low-SoR-density work (high substitutability).} In research, analytics, and content workflows, the ``record'' is mostly documents and versioned artifacts. Execution providers are easily substituted because state is shallow and outputs are highly reviewable.
\paragraph{Limited impact (SoR and invariants dominate).} For core ledgers and high-integrity transaction processing (e.g., settlement), correctness is defined by strict invariants, regulated retention, and deterministic reconciliation. Agents may assist at the margins, but the architecture remains SoR-centered unless the underlying data infrastructure transitions to a federated, zero-copy model \cite{apache2024iceberg} where agents maintain cross-ledger semantic consistency.
\subsection{Empirical protocol}
The predictions in Section \ref{sec:falsifiable} can be evaluated using production telemetry from multi-agent deployments. For Prediction 1, the relevant measurement is the engineering time required to integrate a new execution provider as a function of current ecosystem size $n$. In a topology-dominated system, this should grow proportionally with $n$; in a protocol-mediated system, it should be approximately constant. This can be estimated via developer effort logs (e.g., commit histories, ticket cycle times) in organizations running both legacy SOA and agentic middleware stacks. For Prediction 2, the relevant measurement is $C/T$---total coordination-related engineering cost divided by task throughput---tracked longitudinally as the agent ecosystem grows. Confounds include changes in task complexity distribution $\{k_t\}$ over time and organizational learning effects. A controlled comparison between organizations adopting protocol-mediated architectures versus those extending bespoke integrations would provide the cleanest test. We leave full empirical validation to future work; the purpose of these predictions is to establish that the model is falsifiable in principle.

\subsection{Early Empirical Signals}
While a formal, large-sample test of the coordination cost model awaits longer-term enterprise data, early signals from production agentic deployments align with the sublinear scaling hypothesis. For instance, integration times for new, highly specific tools using the Model Context Protocol (MCP) frequently approach $\mathcal{O}(1)$ developer effort compared to building bespoke API clients \cite{anthropic2024mcp}. Furthermore, the ability to swap specialized agent backends (e.g., changing a search provider or a drafting model) without altering the user-facing intent surface provides anecdotal confirmation that the hourglass waist effectively isolates the topology of the execution layer from the throughput of user requests.
\section{The Execution Layer: Vertical Intelligence}
\label{sec:execution}

When coordination scales sublinearly, the boundary of the firm shifts. As external coordination becomes cheaper than maintaining broad internal capabilities, execution rationally migrates outward into specialized agents. Modularity ceases to be an architectural preference and becomes an economic sorting mechanism.

Historically, SOA decomposed software without decomposing ownership. Agentic systems differ because inference-driven coordination lowers the cost of integrating independent execution units. As integration frictions decline, the efficient unit of production becomes narrower.

\subsection{Vertical Agents and Workflow Ownership}

Vertical agents are AI-enabled services that specialize deeply within a single workflow. They do not span many business functions; they integrate tightly within one. Unlike horizontal suites, which generalize across domains, vertical agents optimize for local objectives, proprietary context, and execution policies.
Their advantage is not the base model. It is the coupling of domain data, evaluation metrics, and tool execution inside a narrow operational loop. As coordination costs fall, specialization deepens.
This is the execution-layer analogue of modular architecture: specialization without integration penalty.

\subsection{Defensibility Through Contextual Accumulation}

A common objection is that commoditized foundation models eliminate durable advantage. This assumes advantage resides in model weights. In vertical systems, it resides in accumulated context.
Enterprise work generates large volumes of unstructured ``dark data'' --- emails, PDFs, chat threads, negotiation traces \cite{idc2012darkdata}. The strategic value of accumulated operational context follows the logic of the Resource-Based View \cite{barney1991firm} but at the data rather than capability level---a distinction developed in the literature on data network effects \cite{martens2020digital} and proprietary training data moats \cite{trask2023data}. A vertical agent embedded in a workflow can capture this data, convert it into evaluation sets and feedback loops, and continuously refine execution policies.
Over time, the system learns why specific contracts fail, which clauses delay negotiation, which client behaviors predict churn. This produces a contextual data network effect: performance improves with embedded use, and replication requires access to proprietary operational traces \cite{chen2017castles}. In the language of the Resource-Based View (RBV), the agent's accumulated context becomes a VRIN (Valuable, Rare, Inimitable, Non-substitutable) asset, whereas the underlying model weights remain a commodity \cite{barney1991firm}.
Crucially, this dynamic reinforces micro-specialization: dark data is highly context-specific (jargon, exception handling, local norms). Generalist systems tend to regress to the mean because they cannot justify the context windows, tuning effort, and workflow instrumentation required to decode a niche vertical's unstructured traces. Recent work on ``task interference'' offers a mechanistic complement: Wang et al. show that training a single model across disparate tasks can induce negative transfer, while mixture-of-experts (MoE) architectures \cite{shazeer2017moe} limit interference by routing inputs to specialized subnetworks \cite{wang2024mome}. The Headless Firm can be read as the organizational analogue of this MoE pattern. Vertical agents can therefore over-optimize for local context and convert dark data into reliable execution.
Finally, this architecture can drive a systemic increase in software quality. In a monolithic suite, features compete for internal resources, often resulting in ``good enough'' mediocrity across the board. In a market of vertical agents, each component must compete on the quality of its specific output to survive. As argued in analyses of modular architectures by Baldwin and Clark, unbundled designs allow individual modules to innovate at their own distinct rates, avoiding the regression-to-the-mean dynamics typical of integrated systems \cite{baldwin2000design}.

\section{The Interaction Layer: Generative Intent}
\label{sec:interaction}

If execution unbundles into vertical agents, interaction must rebundle at the point of use. Specialization lowers coordination cost between machines, but it raises coordination cost for humans: as the execution layer fragments, the user-facing surface area expands. Few users can productively orchestrate dozens of agent-specific interfaces without reintroducing friction at the boundary.

Generative UI is the architectural response to this coordination overhead induced by UI surface fragmentation. The intent layer does not merely ``chat''; it utilizes structured primitives to compile user intent into a constrained interaction plan, orchestrate minimal surfaces to resolve missing parameters, and render auditable confirmations for irreversible actions. The objective is not maximal flexibility but \emph{bounded malleability}: a stable interaction frame within which task-specific micro-interfaces can be composed and discarded.

\subsection{Bounded Malleability and Abstract Primitives}

An engineered intent layer must preserve coherence under heterogeneity. As tasks vary across domains and agents, the interface adapts---but not arbitrarily. To achieve this, Generative UI relies on four abstract primitives:
\begin{enumerate}
    \item \textbf{Intent Compiler:} Parses natural language into a structured, parameterized interaction plan.
    \item \textbf{Constraint Validator:} Checks proposed actions against predefined policy rails and enterprise governance rules before execution.
    \item \textbf{Reversible Preview:} Renders the expected state transition (e.g., a "dry run" of a database write) in a legible format.
    \item \textbf{Policy Surface:} Exposes the underlying permissions and auditing logs contextually to the user.
\end{enumerate}

Persistent anchors provide continuity: visible state, permissions, activity history, and a legible preview of ``what will happen.'' Around these anchors, the system synthesizes ephemeral widgets only when needed to clarify ambiguity or collect missing inputs.
Intent, in this architecture, is compiled rather than loosely interpreted. Natural-language goals are transformed into structured execution artifacts that downstream orchestration can validate and constrain. Where intent is underspecified, the system requests only the information necessary to produce a safe plan. Where actions are irreversible, it presents a confirmation surface that exposes scope, affected systems, and expected state transitions before execution proceeds.
The interaction layer therefore becomes a coordination interface rather than a conversational veneer. It rebundles experience while preserving modular execution.

\subsection{Trust Semantics and Coordination Scalability}

Because agentic execution is stochastic and toolchains can degrade, the interaction layer must make delegation inspectable. Users must be able to see what systems will be touched, what records will change, and which policy context governs the action. Intermediate states, partial completions, and exceptions must be rendered explicitly rather than hidden behind conversational abstraction.
In this sense, Generative UI functions as a contract boundary between intent and execution. It binds user goals to downstream tool invocation under explicit constraints, transforming open-ended language into legible state transitions.
Crucially, this layer must itself be evaluated as a coordination mechanism. The relevant question is not aesthetic quality but structural scalability: does it reduce clarification cycles as execution complexity grows? Does it keep irreversible actions legible under variation? Does it remain stable under tool-schema change and policy drift? If interface variability scales with ecosystem diversity, cognitive load becomes topology-dependent and the hourglass collapses from above.
Recent work on generative interfaces and agentic HCI \cite{leviathan2025generative, wang2025agentic} motivates these architectural primitives, while advances in dynamic interface synthesis and test-time adaptation suggest that interaction can be evaluated systematically rather than treated as an aesthetic layer \cite{chen2025genui, xia2024generative}.
The interaction layer thus mirrors the logic of the execution layer. A complete evaluation framework for the generative intent layer would require at minimum three metrics: (1) clarification cycle reduction rate---the ratio of task completions requiring zero or one clarification exchange versus baseline unaugmented interface; (2) irreversible action legibility---user accuracy in predicting post-action state under time pressure; (3) cognitive load stability under tool-schema drift---whether error rates remain bounded as the underlying agent ecosystem evolves. We leave empirical measurement of these metrics to future work but propose them as the minimal sufficient evaluation protocol for any production deployment of generative intent architectures. Just as protocol standardization compresses integration edges among agents, generative intent compresses interaction edges for users---provided that bounded malleability and explicit confirmation semantics prevent cognitive load from reintroducing topology-driven coordination costs.

\section{The Hourglass Architecture of the ``Headless'' Firm}

The preceding sections derived two structural shifts. 
First, execution unbundles into a competitive market of vertical agents once coordination scales sublinearly. 
Second, interaction must rebundle through generative intent to prevent fragmentation at the user boundary. 

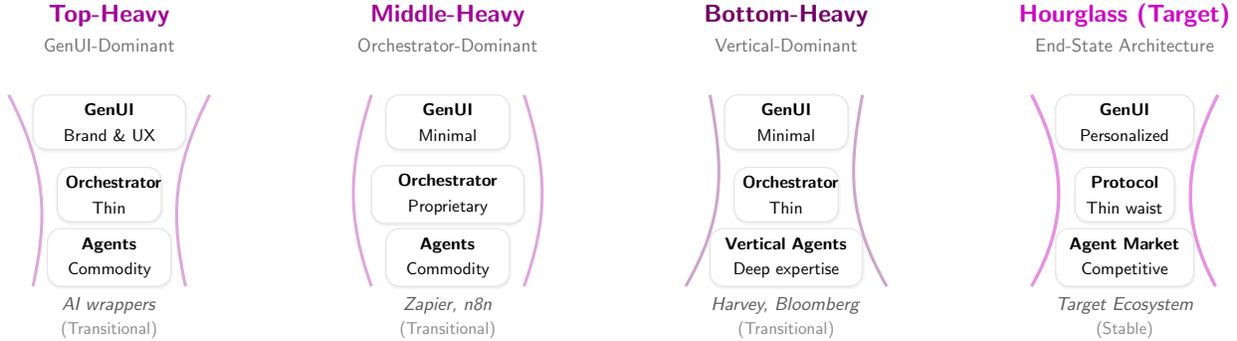
\begin{figure}[!htbp]
  \centering
  \resizebox{\textwidth}{!}{
  \begin{tikzpicture}[
    node distance=1.0cm,
    font=\sffamily
  ]
    
    \begin{scope}[xshift=0cm, yshift=0cm]
      \node[font=\sffamily\bfseries\Large, labBlue] at (0, 6.0) {Top-Heavy};
      \node[font=\sffamily\normalsize, labDark!60] at (0, 5.4) {GenUI-Dominant};
      
      \node[labNode, text width=2.8cm, align=center, minimum height=1.1cm, inner sep=5pt] (top1) at (0, 3.8) {
        \textbf{GenUI} \\[3pt]
        {\small Brand \& UX}
      };
      \node[labNode, text width=1.8cm, align=center, minimum height=0.7cm, inner sep=5pt] (mid1) at (0, 2.3) {
        \textbf{Orchestrator} \\[3pt]
        {\small Thin}
      };
      \node[labNode, text width=2.2cm, align=center, minimum height=0.7cm, inner sep=5pt] (bot1) at (0, 1.0) {
        \textbf{Agents} \\[3pt]
        {\small Commodity}
      };
      
      \draw[draw=labBlue, line width=1.8pt, opacity=0.35, rounded corners=12pt] 
        ($(top1.north west)+(-0.5, 0)$) .. controls ($(mid1.west)+(-0.2, 0.5)$) and ($(mid1.west)+(-0.2, -0.5)$) .. ($(bot1.south west)+(-0.3, 0)$);
      \draw[draw=labBlue, line width=1.8pt, opacity=0.35, rounded corners=12pt] 
        ($(top1.north east)+(0.5, 0)$) .. controls ($(mid1.east)+(0.2, 0.5)$) and ($(mid1.east)+(0.2, -0.5)$) .. ($(bot1.south east)+(0.3, 0)$);
      
      \node[font=\sffamily\normalsize, labDark!70, align=center] at (0, 0.0) {\textit{AI wrappers}};
      \node[font=\sffamily\small, labDark!50] at (0, -0.5) {(Transitional)};
    \end{scope}
    
    \begin{scope}[xshift=7.0cm, yshift=0cm]
      \node[font=\sffamily\bfseries\Large, labPurple] at (0, 6.0) {Middle-Heavy};
      \node[font=\sffamily\normalsize, labDark!60] at (0, 5.4) {Orchestrator-Dominant};
      
      \node[labNode, text width=2.2cm, align=center, minimum height=0.7cm, inner sep=5pt] (top2) at (0, 3.8) {
        \textbf{GenUI} \\[3pt]
        {\small Minimal}
      };
      \node[labNode, text width=2.8cm, align=center, minimum height=1.1cm, inner sep=5pt] (mid2) at (0, 2.3) {
        \textbf{Orchestrator} \\[3pt]
        {\small Proprietary}
      };
      \node[labNode, text width=2.2cm, align=center, minimum height=0.7cm, inner sep=5pt] (bot2) at (0, 1.0) {
        \textbf{Agents} \\[3pt]
        {\small Commodity}
      };
      
      \draw[draw=labPurple, line width=1.8pt, opacity=0.35, rounded corners=12pt] 
        ($(top2.north west)+(-0.3, 0)$) .. controls ($(mid2.west)+(-0.5, 0.5)$) and ($(mid2.west)+(-0.5, -0.5)$) .. ($(bot2.south west)+(-0.3, 0)$);
      \draw[draw=labPurple, line width=1.8pt, opacity=0.35, rounded corners=12pt] 
        ($(top2.north east)+(0.3, 0)$) .. controls ($(mid2.east)+(0.5, 0.5)$) and ($(mid2.east)+(0.5, -0.5)$) .. ($(bot2.south east)+(0.3, 0)$);
      
      \node[font=\sffamily\normalsize, labDark!70, align=center] at (0, 0.0) {\textit{Zapier, n8n}};
      \node[font=\sffamily\small, labDark!50] at (0, -0.5) {(Transitional)};
    \end{scope}
    
    \begin{scope}[xshift=14.0cm, yshift=0cm]
      \node[font=\sffamily\bfseries\Large, labAccent] at (0, 6.0) {Bottom-Heavy};
      \node[font=\sffamily\normalsize, labDark!60] at (0, 5.4) {Vertical-Dominant};
      
      \node[labNode, text width=2.2cm, align=center, minimum height=0.7cm, inner sep=5pt] (top3) at (0, 3.8) {
        \textbf{GenUI} \\[3pt]
        {\small Minimal}
      };
      \node[labNode, text width=1.8cm, align=center, minimum height=0.7cm, inner sep=5pt] (mid3) at (0, 2.3) {
        \textbf{Orchestrator} \\[3pt]
        {\small Thin}
      };
      \node[labNode, text width=2.8cm, align=center, minimum height=1.1cm, inner sep=5pt] (bot3) at (0, 1.0) {
        \textbf{Vertical Agents} \\[3pt]
        {\small Deep expertise}
      };
      
      \draw[draw=labAccent, line width=1.8pt, opacity=0.35, rounded corners=12pt] 
        ($(top3.north west)+(-0.3, 0)$) .. controls ($(mid3.west)+(-0.2, 0.5)$) and ($(mid3.west)+(-0.2, -0.5)$) .. ($(bot3.south west)+(-0.5, 0)$);
      \draw[draw=labAccent, line width=1.8pt, opacity=0.35, rounded corners=12pt] 
        ($(top3.north east)+(0.3, 0)$) .. controls ($(mid3.east)+(0.2, 0.5)$) and ($(mid3.east)+(0.2, -0.5)$) .. ($(bot3.south east)+(0.5, 0)$);
      
      \node[font=\sffamily\normalsize, labDark!70, align=center] at (0, 0.0) {\textit{Harvey, Bloomberg}};
      \node[font=\sffamily\small, labDark!50] at (0, -0.5) {(Transitional)};
    \end{scope}
    
    \begin{scope}[xshift=21.0cm, yshift=0cm]
      \node[font=\sffamily\bfseries\Large, labCyan] at (0, 6.0) {Hourglass (Target)};
      \node[font=\sffamily\normalsize, labDark!60] at (0, 5.4) {End-State Architecture};
      
      \node[labNode, text width=2.5cm, align=center, minimum height=0.9cm, inner sep=5pt] (top4) at (0, 3.8) {
        \textbf{GenUI} \\[3pt]
        {\small Personalized}
      };
      \node[labNode, text width=1.7cm, align=center, minimum height=0.8cm, inner sep=5pt] (mid4) at (0, 2.3) {
        \textbf{Protocol} \\[3pt]
        {\small Thin waist}
      };
      \node[labNode, text width=2.5cm, align=center, minimum height=0.9cm, inner sep=5pt] (bot4) at (0, 1.0) {
        \textbf{Agent Market} \\[3pt]
        {\small Competitive}
      };
      
      \draw[draw=labCyan, line width=2.0pt, opacity=0.45, rounded corners=12pt] 
        ($(top4.north west)+(-0.5, 0)$) .. controls ($(mid4.west)+(-0.15, 0.6)$) and ($(mid4.west)+(-0.15, -0.6)$) .. ($(bot4.south west)+(-0.5, 0)$);
      \draw[draw=labCyan, line width=2.0pt, opacity=0.45, rounded corners=12pt] 
        ($(top4.north east)+(0.5, 0)$) .. controls ($(mid4.east)+(0.15, 0.6)$) and ($(mid4.east)+(0.15, -0.6)$) .. ($(bot4.south east)+(0.45, 0)$);
      
      \node[font=\sffamily\normalsize, labDark!70, align=center] at (0, 0.0) {\textit{Target Ecosystem}};
      \node[font=\sffamily\small, labDark!50] at (0, -0.5) {(Stable)};
    \end{scope}
    
  \end{tikzpicture}
  }
\caption{Convergence toward the hourglass equilibrium. Transitional configurations concentrate value in a single architectural layer (top-heavy: brand and UX; middle-heavy: orchestration logic; bottom-heavy: domain expertise). The stable end-state emerges when protocol standardization commoditizes orchestration, pushing differentiation toward the high-context edges: personalized intent above and deep domain execution below. This architectural evolution mirrors the layered convergence of Internet protocol stacks documented by Akhshabi and Dovrolis \cite{akhshabi2011evolution}. The stability condition --- that the protocol waist remain thin --- is formalized as the condition $\gamma < 1$ in Section \ref{sec:model}.}
\label{fig:hourglass}
\end{figure}

Taken together, these forces imply a specific architectural equilibrium. 
Execution fragments downward; interaction coheres upward; coordination compresses in between. 
The resulting structure is an hourglass, shown in Figure~\ref{fig:hourglass}.

Wide specialization at the bottom and wide personalization at the top are connected by a thin protocol waist that standardizes coordination.

Adapting the layered protocol model of \cite{akhshabi2011evolution}, the hourglass consists of three layers:
\paragraph{Top (Generative UI) — Solves the coordination overhead induced by UI surface fragmentation.} A dynamic interaction layer captures user intent and synthesizes ephemeral interfaces. Rather than navigating dozens of specialized agents, users interact with a unified intent surface. Interaction is rebundled even as execution is unbundled.
\paragraph{Middle (Protocol Orchestrator) — Solves the verification and governance overhead of delegated execution.} A thin protocol layer compiles intent into structured commands and enforces safety, provenance, and policy-gated writes. Unlike a passive gateway, the orchestrator maintains a consistent view of state despite stochastic agent behavior \cite{mei2025aios}. Coordination is compressed into standardized contracts.
\paragraph{Bottom (Vertical Agent Market) — Solves the overhead of maintaining internal capabilities in high-velocity knowledge domains.} A competitive ecosystem of micro-specialized agents executes tasks. Because these agents are decoupled from UI and orchestration logic, they can evolve at the speed of their domain without destabilizing the broader system.
The architecture is structurally asymmetric. 
Differentiation concentrates at the edges—personalized interaction above and deep domain expertise below—while the middle collapses into a thin, standardized protocol waist. 
As Figure~\ref{fig:hourglass} illustrates, transitional configurations concentrate value in a single layer (top-heavy, middle-heavy, or bottom-heavy). The stable end-state emerges when protocol standardization commoditizes orchestration and competitive agent markets commoditize execution, forcing durable advantage toward the high-context edges.
The ``Headless'' Firm is therefore not defined by the absence of structure, but by the separation of interaction and execution through a thin coordination layer.

\subsection{Transitional Patterns and Convergence}

Current systems exhibit unstable intermediate configurations in which value concentrates in a single layer.
\emph{Top-heavy} systems (e.g., AI wrappers) concentrate value in brand and UX, while execution relies on commodity APIs.  
\emph{Middle-heavy} systems (e.g., proprietary orchestrators such as Zapier) concentrate value in coordination logic.  
\emph{Bottom-heavy} systems (e.g., vertical AI specialists) concentrate value in domain expertise and proprietary data.
These are transitional equilibria. As protocol standards mature and agent markets deepen, coordination logic commoditizes and execution becomes interchangeable. Differentiation migrates to the high-context edges: personalized interaction at the top and deep specialization at the bottom.
The hourglass is therefore not a design preference but a convergence outcome.

\subsection{Coordination scalability: the manager-to-managed inversion}
The hourglass architecture is not merely a technical design pattern---it represents a fundamental inversion of the traditional management pyramid. The hourglass inverts the relationship between hierarchical control and transaction costs (as discussed in Section 2). Instead of imperative micromanagement (``execute task X, then Y, then report status''), the orchestrator operates declaratively: it routes intent to capable agents, enforces policy constraints, and verifies outcomes. The orchestrator does not coordinate \emph{how} agents execute---only \emph{what} they must satisfy (safety, provenance, state consistency). This shifts the coordination model from hierarchical command-and-control to market-based outcome verification \cite{malone1994interdisciplinary}.
First, standardized protocols eliminate bespoke integration work. Unlike microservices that require custom API contracts and versioning coordination, agents in a mature hourglass ecosystem communicate through shared protocols (e.g., Model Context Protocol, standardized tool schemas), reducing the orchestrator's role to routing and validation rather than negotiation.
Second, outcome-based evaluation enables autonomy. The orchestrator does not monitor \emph{process}---it verifies \emph{results}. An agent that consistently produces correct outputs requires no supervision, analogous to the shift from time-based employment to task-based completion.
Third, market mechanisms replace top-down allocation. In a competitive agent marketplace, tasks are allocated via bidding, reputation, or capability matching rather than centralized assignment \cite{roth2002economist,shahidi2025coasean}---the orchestrator becomes an auctioneer, not a planner, aligning with Hayek's argument that decentralized signals aggregate information more efficiently than central planning \cite{hayek1945knowledge}.
Finally, coordination shifts from static routing to intelligent delegation \cite{tomasev2026intelligent}: the dynamic transfer of authority based on real-time context, where the orchestrator negotiates delegation contracts that enforce auditability and the right to revoke authority if safety constraints are violated.
This inversion has profound implications for scalability. If coordination overhead grows sublinearly (or even logarithmically) with the number of agents, then the economic ceiling on firm size shifts dramatically. A single orchestrator could theoretically manage millions of micro-specialized agents---not through heroic management, but through \emph{protocol-mediated indifference}. The orchestrator does not care \emph{which} legal research agent drafts a memo, only that the output meets specified quality thresholds and provenance requirements. Empirically, this predicts a bifurcation in the firm size distribution (recall Figure 1): a small number of massive orchestrators (protocol-setters like AWS, Stripe, or future MCP-compliant platforms) and a long tail of hyper-specialized execution agents. The middle---integrated firms that bundle orchestration with execution---becomes economically unstable, squeezed between the efficiency of thin protocols and the depth of vertical specialists. This mirrors the platform economics literature's prediction that two-sided markets concentrate around dominant intermediaries while enabling a long tail of niche suppliers \cite{parker2016platform}. This is why the hourglass is not just architecturally elegant but economically inevitable: it is the only structure that scales coordination costs sublinearly while preserving the adaptability benefits of decentralized specialization. The transitional patterns (top-heavy, middle-heavy, bottom-heavy) represent temporary market inefficiencies that will erode as protocol standardization and agent marketplaces mature.

\section{Economic and Labor Implications}
\label{sec:implications}

If the hourglass architecture is the equilibrium form under sublinear coordination scaling, then firm boundaries and software economics must adjust accordingly. We analyze three specific shifts: the dual effect of AI on firm size, the rise of the "scalable boutique," and the inversion of the make-or-buy decision.

\subsection{The Dual Effect on Firm Size}
The preceding analysis raises an apparent paradox: if AI reduces coordination costs, why predict smaller firms rather than larger ones? Williamson's 1967 model of hierarchical control suggests that reducing ``loss of control'' through better monitoring should enable firms to grow \cite{williamson1967hierarchical}. In his framework, firm size is limited by the degradation of information as it passes through hierarchical layers---formalized as a control transmission factor $\beta < 1$, where each additional layer multiplies the control loss. If AI increases $\beta$ (by reducing information distortion through real-time dashboards and automated reporting) and expands the span of control $s$ (by automating routine supervision), then the optimal number of hierarchical levels $n$ could increase, supporting larger integrated firms.

Calvo and Wellisz (1978) refined this model, arguing that the nature of supervision matters: if employees cannot predict when they are being monitored (e.g., through stochastic AI-driven audits), the loss of control may be even lower than Williamson's deterministic model suggests \cite{calvo1978supervision}. This implies AI could enable \emph{dramatically} larger hierarchical firms by solving the principal-agent problem at scale. Empirical evidence supports this mechanism: Brynjolfsson et al. (2023) find that generative AI tools increase worker productivity by 14\% on average (and 34\% for novice workers) by disseminating best practices and reducing information asymmetry \cite{brynjolfsson2023generative}. Murray et al. (2021) document how algorithmic management expands the span of control by automating routine supervision, enabling managers to oversee larger teams without degradation in oversight quality \cite{murray2021algorithms}. Puranam et al. (2014) observe that digital communication technologies enable flatter organizational structures by reducing the need for middle management layers \cite{puranam2014new}. In the extreme case, decentralized autonomous organizations (DAOs) replace hierarchical control entirely with smart contracts, theoretically achieving $\beta \approx 1$ through protocol-enforced transparency \cite{hassan2021daos}.

However, this analysis assumes a critical constraint: that the \emph{knowledge embodied in the firm's processes is stable}. Williamson's model treats the firm's production function as fixed; the challenge is purely one of coordination and monitoring. In this regime, AI's reduction of internal coordination costs unambiguously favors larger integrated firms, as the benefits of economies of scale and scope compound without the offsetting cost of constant adaptation.

The hourglass architecture emerges when we relax this assumption. In high-velocity knowledge domains---medicine, law, compliance, software engineering---the firm faces a \emph{temporal constraint} that Williamson's model does not capture: the decay rate of codified knowledge \cite{jones2009burden}. Internal tooling, standard operating procedures, and domain-specific models become obsolete faster than they can be maintained. This introduces a ``complexity wall'': the cost of continuously rewriting internal systems to track the frontier of knowledge scales superlinearly with firm size, because larger firms have more legacy systems, more interdependencies, and more organizational inertia \cite{henderson1990architectural, tainter1988collapse}.
AI's effect on firm size is therefore \emph{asymmetric}. It reduces \emph{both} internal coordination costs (higher $\beta$, larger $s$, fewer $n$) \emph{and} external transaction costs (cheaper market-based coordination via agent marketplaces and standardized protocols). The net effect depends on which dominates. We argue that in high-velocity domains, the external transaction cost reduction dominates for four reasons. First, the firm's internal AI investment faces a compounding disadvantage: foundation model capabilities advance faster than any single firm can fine-tune against, meaning internal AI tooling depreciates toward the frontier at an accelerating rate. External specialists who serve many clients can amortize continuous retraining and evaluation costs; integrated firms cannot. This is a specific instance of the general principle that adaptation costs scale with the breadth of the capability portfolio maintained internally \cite{henderson1990architectural}. Second, a vertical specialist (e.g., a legal research agent) can amortize the cost of tracking regulatory changes, updating domain models, and refining evaluation harnesses across hundreds of clients---an integrated firm must bear this cost internally for each domain it internalizes, creating diseconomies of scope in adaptation. Third, external specialists can update their models and processes independently, without coordinating release cycles or regression-testing against the broader system; even an AI-optimized monolith must centrally plan updates, creating bottlenecks \cite{hayek1945knowledge}. Fourth, in a market of vertical agents, poor performers are replaced instantly thanks to low switching costs---in an integrated firm, underperforming internal teams face organizational friction (politics, sunk costs, career concerns) that slows the feedback loop.
Empirically, this predicts a \emph{conditional} effect: AI will enable larger firms in domains with stable knowledge (e.g., logistics, manufacturing, routine transaction processing), where Williamson's mechanism dominates. But in high-velocity domains, AI will accelerate unbundling, as the transaction cost collapse outpaces the control loss reduction. The hourglass is the equilibrium architecture for the latter regime.
This resolves the paradox: AI does not uniformly shrink or grow firms. It \emph{bifurcates} the firm size distribution based on the knowledge decay rate of the domain. Slow-moving domains consolidate; fast-moving domains fragment. The Great Unbundling is domain-specific, not universal.

\subsection{The Scalable Boutique: Return on Talent vs. Return on Scale}
This bifurcation carries an asymmetric wage implication that operates primarily in fast-moving domains: precisely where unbundling dominates, small AI-augmented firms can pay \emph{higher} salaries than large integrated incumbents. However, this dynamic is strictly conditional on surplus capture: scalable boutiques can only pay a wage premium if they capture the surplus generated by AI leverage at the orchestration and specialization layers, rather than having it extracted via platform rents or inference pricing by foundation model providers. Assuming this condition holds, while in slow-moving domains where consolidation prevails the traditional large-firm wage premium likely persists, empirical evidence suggests this premium is broadly eroding, and in some high-tech domains, reversing entirely.

The mechanism is \emph{marginal productivity of labor}. In a traditional large firm, an individual employee's contribution is diluted across thousands of colleagues and layers of bureaucracy. In a small, AI-augmented firm, each employee controls a disproportionate ``leverage'' of capital and technology. A lean startup using generative AI and cloud infrastructure can serve millions of users (analogous to pre-agentic high-leverage cases such as WhatsApp, which served 450 million users with 55 engineers at acquisition\footnote{WhatsApp had approximately 55 employees at the time of its \$19B acquisition by Facebook in February 2014, serving over 450 million monthly active users. See contemporaneous reporting: e.g., Olga Khazan, \textit{The Atlantic}, February 19, 2014.}---a ratio that agentic architectures may extend further by replacing human coordination with protocol-mediated agent delegation), generating revenue per employee that rivals or exceeds that of 100,000-person conglomerates. Autor et al. (2020) document the rise of ``superstar firms'' that achieve extraordinary productivity with lean headcounts, capturing market share through technological advantage rather than scale \cite{autor2020superstar}.

Song et al. (2019) provide complementary evidence: two-thirds of the rise in U.S. earnings inequality from 1978 to 2013 occurred \emph{between} firms, not within them \cite{song2019firming}. High-wage workers increasingly sort into high-wage firms, and these high-wage firms are not necessarily the largest---they are the most productive. This sorting is enabled by technology: cloud platforms, AI tools, and global logistics allow small firms to access the same infrastructure once reserved for giants, leveling the playing field in operational efficiency while preserving advantages in adaptability.

The hourglass architecture amplifies this dynamic. In a headless firm, a small team can orchestrate a vast network of vertical agents without bearing the overhead of a large internal workforce---creating a ``scalable boutique'' model that captures enterprise revenue while maintaining startup agility. Three mechanisms drive the higher compensation potential:
\begin{enumerate}\setlength{\itemsep}{2pt}
    \item \textbf{Hyper-leveraged productivity.} A lean firm (e.g., Midjourney generating an estimated \$200M in revenue with fewer than 40 employees) achieves revenue per employee an order of magnitude above a traditional software company; even a modest marginal product payout vastly exceeds a large-firm median salary.
    \item \textbf{Catastrophic retention risk.} Small firms face a ``must-win'' scenario for talent: losing a single key engineer is catastrophic at 10 people but negligible at 10,000, forcing above-market retention wages.
    \item \textbf{Open market competition.} Unlike large firms that exercise monopsony power over employees in similar roles, small firms must compete in the open market for every hire, bidding wages upward.
\end{enumerate}

Cobb and Lin (2017) document this shift empirically, showing that the large-firm wage premium has declined significantly since the 1980s, particularly in technology-intensive industries \cite{cobb2017growing}---the same premium that historically justified sorting into large firms has been structurally eroded by the technology leverage available to small teams. This trend accelerates in the agentic era: as orchestration becomes commoditized and execution is delegated to external agents, the optimal firm size for maximizing \emph{return on talent} (not return on scale) shrinks dramatically.

This has profound implications for labor markets. The Great Unbundling does not merely redistribute work from large firms to small ones---it redistributes \emph{value capture} from shareholders to high-skilled workers. In a world where a 10-person firm can generate \$100M in revenue, the bargaining power of each employee is immense. This predicts a bifurcation in compensation: a small number of highly skilled workers in scalable boutiques will capture extraordinary wages, while workers in commoditized roles (whether in large firms or as external agents) face downward pressure from automation and competition.

\subsection{The Inversion of the Make-or-Buy Decision}

\begin{figure}[ht]
  \centering
  \begin{subfigure}[b]{0.48\textwidth}
    \centering
    \begin{tikzpicture}[font=\sffamily\small, scale=0.6, transform shape]
      \draw[labAxis] (0,0) -- (10,0) node[right, labDark!60] {scale / usage};
      \draw[labAxis] (0,0) -- (0,4.6) node[above, labDark!60] {marginal cost};

      \draw[labLegacy] (0.5,0.9) -- (9.5,0.9) node[right, labDark!50] {legacy SaaS};

      \fill[labFill] (0.5,0.9) -- (3,1.6) -- (6,2.7) -- (9.5,3.9) -- (9.5,0.9) -- cycle;
      \draw[labData] plot[smooth] coordinates {(0.5,0.9) (3,1.6) (6,2.7) (9.5,3.9)};
      
      \node[labBlue, font=\sffamily\bfseries, anchor=south west] at (4.0,3.5) {Agentic AI (Inference)};

      \node[labDark!70, align=left, font=\sffamily\tiny] at (6.0,1.8) {Subscription ceiling;\\Price for outcomes};
    \end{tikzpicture}
    \caption{Marginal cost dynamics.}
    \label{fig:marginal_cost}
  \end{subfigure}
  \hfill
  \begin{subfigure}[b]{0.48\textwidth}
    \centering
    \begin{tikzpicture}[font=\sffamily\small, scale=0.6, transform shape]
      \draw[labAxis] (0,0) -- (10,0) node[right, labDark!60] {time};
      \draw[labAxis] (0,0) -- (0,4.6) node[above, labDark!60] {innovation rate / quality};

      \draw[labLegacy] plot[smooth] coordinates {(0.5,0.7) (2,1.1) (4,1.4) (6,1.55) (8,1.65) (9.5,1.7)};
      \node[labDark!40, anchor=north] at (5.5,1.4) {monolith (coupled)};

      \draw[labData] plot[smooth] coordinates {(0.5,0.7) (2,1.4) (4,2.2) (6,3.0) (8,3.7) (9.5,4.1)};
      \node[labBlue, font=\sffamily\bfseries, anchor=south east] at (9.0,3.9) {Modular};

      \begin{scope}
        \clip (0.2, 0) rectangle (9.5, 5);
        \fill[labCyan!20, opacity=0.4] 
          (0.5,0.7) plot[smooth] coordinates {(0.5,0.7) (2,1.4) (4,2.2) (6,3.0) (8,3.7) (9.5,4.1)} --
          (9.5,1.7) -- 
          plot[smooth] coordinates {(9.5,1.7) (8,1.65) (6,1.55) (4,1.4) (2,1.1) (0.5,0.7)} -- cycle;
      \end{scope}

      \draw[labArrow, draw=labCyan, dashed] (8,1.7) -- (8,3.6) node[midway, left=2pt, labCyan, font=\sffamily\bfseries\tiny, align=right] {agility\\gap};

    \end{tikzpicture}
    \caption{Innovation rate divergence.}
    \label{fig:innovation_rate}
  \end{subfigure}
  \caption{Structural economic shifts in the agentic era. (a) Stylized representation of marginal cost dynamics: traditional SaaS approaches zero marginal cost at scale while agentic AI inference incurs per-token cost that scales with usage volume, following empirical inference cost curves documented in \cite{arxiv2025inference}. (b) Conceptual illustration of innovation rate divergence between monolithic and modular architectures, consistent with modularity theory \cite{baldwin2000design} but not derived from empirical data. Both panels are schematic; empirical calibration is left to future work.}
  \label{fig:combined_economics}
\end{figure}
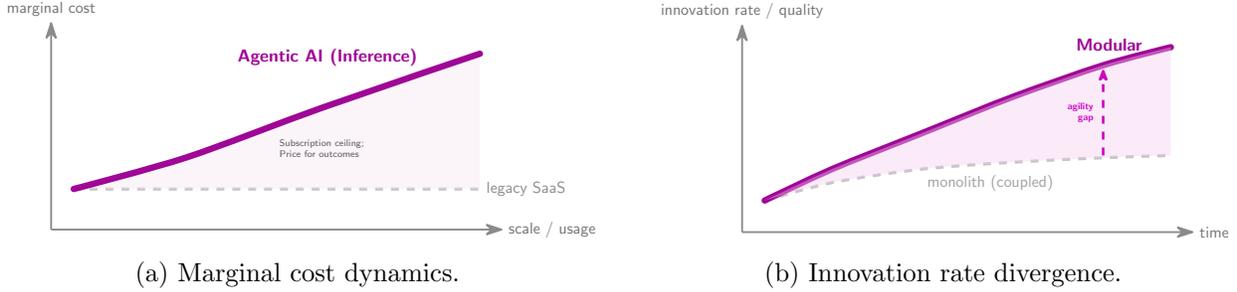

The economic consequences of the hourglass equilibrium are summarized in Figure~\ref{fig:combined_economics}. 
Panel (a) shows how inference-based systems undermine the traditional zero-marginal-cost logic of software, while Panel (b) illustrates how modular agentic ecosystems outpace tightly coupled monoliths through independent, compounding innovation.

Historically, buying SaaS was cheaper than hiring engineers to build and maintain bespoke internal tools. Agentic AI can invert this equation.

First, many enterprise offerings package general-purpose models behind seat-based pricing. In practice, customers may pay for integration, governance, and procurement convenience---but they may also face perceived ``per-token'' markups when model usage is embedded in premium stock-keeping units (SKUs).

Second, AI can weaken the classic ``zero marginal cost'' intuition of software \cite{rifkin2014zero}. While traditional SaaS distributes a fixed codebase at low incremental cost, AI features incur ongoing inference cost \emph{per interaction}, pushing vendors toward usage-based pricing and making high-frequency workflows expensive at scale \cite{aiindex2025}. This pressure is one reason seat-based pricing feels brittle once a product becomes a thin wrapper around metered model calls.

Third, AI coding assistants and agentic development workflows reduce the time-to-build for narrow tools. This enables \emph{disposable software} \cite{thomas2002disposable}: lightweight, purpose-built scripts or apps created quickly for a temporary need and discarded or rewritten as requirements change, rather than committing to a complex long-lived platform. Crucially, this lowers the barrier to exit as much as the barrier to entry. Because the sunk cost of these ephemeral agents is near-zero, firms can switch providers instantly without the ``lock-in'' friction and high switching costs that classically define legacy enterprise software markets \cite{shapiro1998information}, forcing agents to compete continuously on performance.

Recent analyses of inference economics \cite{arxiv2025inference} indicate that while training costs are one-time capital expenditures (CapEx), inference costs behave as variable operating expenses (OpEx) that scale linearly with usage. Unlike legacy SaaS, where the marginal cost of serving the $N^{\text{th}}$ user approaches zero, the marginal cost of an agentic workflow remains significant. This pushes pricing away from flat-rate subscriptions (which bleed margin under heavy use) toward outcome-based or consumption-based models.

This shift marks an economic transition from ``SaaS'' (selling tools to humans) to ``Service-as-Software'' (selling outcomes to firms). While traditional SaaS competes for a relatively constrained IT budget, vertical agents compete for the much larger payroll and outsourced-services budget. As argued by HFS Research \cite{venkatesan2025services}, this creates a non-linear growth opportunity by expanding the addressable market from software spend to services spend.

\subsection{The Democratization of the Stack}

The hourglass architecture creates two symmetric entry points (Figure~\ref{fig:hourglass}). At the \textbf{frontend entry}, a small team can build a compelling generative UI by orchestrating best-in-class agents rather than recreating backend capabilities. At the \textbf{backend entry}, a domain expert can build a vertical agent by encoding specialized process knowledge, relying on the ecosystem for distribution and interaction.

Because coordination is standardized at the waist, neither side must internalize the entire stack. This operationalizes a strategic form of Inverse Conway's Law \cite{skelton2019team}: firms can adopt a headless architecture to induce modular, high-velocity organizational structure.

This aligns with the ``Software 2.0'' paradigm \cite{karpathy2017software2}. Scarcity shifts from writing explicit code toward curating data, defining objectives, and constructing evaluation harnesses. In a headless ecosystem, builders compete on semantics rather than syntax.

\section{Open Problems}

The central open question is whether the relocated coordination layer---verification, governance, and market allocation---can scale with throughput rather than topology in practice. Each item below represents a mechanism that could tip the balance toward re-centralization if left unsolved.

\paragraph{Incentive design and market structure.} Agent marketplaces require mechanisms for reputation, routing, pricing, and capability discovery. The open question is how to prevent race-to-the-bottom equilibria while preserving meaningful competition and specialization.

\paragraph{Verification, liability, and auditability.} As execution is delegated to autonomous agents, the primary friction shifts from vendor selection to behavioral verification. This motivates research in evaluation harnesses, persistent agent identity, decentralized identity (DID) systems \cite{w3c2022did}, provenance tracking, and liability allocation under stochastic execution.

\paragraph{Protocol standardization.} The thin waist depends on shared tool contracts, semantic interoperability, and agent-interface specifications (e.g., Model Context Protocol \cite{anthropic2024mcp}). Without standards, coordination costs re-fragment into bespoke integrations.

\paragraph{Data governance and vertical integration.} Vertical agents accumulate high-context operational traces. Ensuring privacy, regulatory compliance, and controlled reuse of proprietary workflow data is foundational for defensible specialization without unacceptable leakage.

\paragraph{State integrity and composable systems of record.} Multi-agent ecosystems must preserve invariants across composable ERPs and zero-copy architectures. Whether agents can safely mediate irreversible writes and high-integrity state transitions remains an open empirical question.

\paragraph{Robustness under distribution shift.} Agent performance can degrade under changing tools, policies, or adversarial inputs. Understanding cascading failures in multi-agent toolchains and designing graceful degradation remains open.

\section{Critical Discussion: The Risks of Re-Centralization}
\label{sec:discussion}

If the hourglass architecture is the equilibrium under sublinear coordination scaling, then its stability depends on preventing coordination costs from re-accumulating elsewhere in the system. Each of the following risks represents a distinct mechanism through which coordination could revert from throughput-based scaling back to topology-based scaling, reintroducing centralization pressures.

\subsection{The Aggregator Paradox: The Risk of a ``Thick Waist''}
\label{sec:thickwaist}

A central critique of the hourglass thesis is that protocol layers in digital markets rarely capture value; instead, aggregators built atop them do, a phenomenon formalized as platform envelopment \cite{eisenmann2006strategies}\footnote{See also Thompson (2015) for an influential practitioner account of aggregation dynamics in digital markets: \url{https://stratechery.com/aggregation-theory/}}. If the Orchestrator controls identity, routing, reputation, or pricing, it may cease to function as a neutral protocol and instead become a rent-extracting gatekeeper — a ``thick waist'' that re-centralizes coordination.

This is not merely a value-capture concern but a scaling concern. A protocol remains thin only if it minimizes discretionary policy and pricing power. Once routing decisions, identity verification, or reputation scoring become proprietary and vertically integrated, coordination cost begins scaling with platform governance rather than standardized contracts.

Open standards such as the Model Context Protocol (MCP) \cite{anthropic2024mcp} reduce this risk by commoditizing tool interaction before monopoly control can emerge. Historically, attempts to build ``smart pipes'' (e.g., ESBs) re-centralized integration logic and ultimately failed because they internalized too much coordination intelligence. As formalized by the classic end-to-end principle in system design \cite{saltzer1984end}, the stable design is the minimal pipe that allows maximal intelligence at the edges. 

However, this outcome is contingent on protocol governance. If standards remain vendor-controlled or allow proprietary extensions, the waist can re-thicken. Sublinear coordination scaling therefore depends not just on technical design, but on institutional neutrality.

\subsection{The Cognitive Load Trap: Interaction Re-Thickening}
\label{sec:cognitiveload}

The hourglass assumes that generative interfaces reduce human coordination costs by compressing navigation into intent. Yet HCI research on Cognitive Load Theory suggests that users rely on spatial consistency and stable affordances to construct mental models \cite{sweller1988cognitive}. A fully malleable interface risks increasing cognitive coordination cost rather than reducing it.
If interface variability scales with task diversity, then human coordination burden may grow with system complexity. In that case, interaction cost re-scales with topology rather than collapsing into an intent abstraction layer.
The viable design path is \emph{bounded malleability}. Generative interfaces must preserve stable anchors (navigation rails, persistent affordances) while synthesizing ephemeral micro-interfaces around tasks. The objective is to eliminate navigation friction without eliminating cognitive predictability. 
If this balance fails, the interaction layer becomes a new coordination bottleneck, undermining the hourglass equilibrium.

\subsection{The Jevons Paradox of Inference: The Return of Marginal Cost}
\label{sec:liability}
Economic theory suggests that as AI inference costs drop (following a Moore's Law trajectory), the pressure to shift from ``SaaS'' (flat-rate) to ``Service-as-Software'' \cite{fersht2025saas} (outcome-based) pricing may evaporate. If intelligence becomes ``too cheap to meter,'' why unbundle?
This view ignores the Jevons Paradox \cite{jevons1865coal}. As the cost per token drops, demand for reasoning depth increases. We are witnessing a structural shift toward test-time compute scaling \cite{snell2024scaling}, where models allocate additional inference cycles to verify and refine chains of thought. Early ``Level 2'' reasoners (like OpenAI's o1) demonstrated that performance scales not just with model parameters, but with ``thinking time''---ensuring that the computational cost of high-integrity judgments remains significant. More precisely, the marginal cost $MC$ of executing a task via an autonomous agent is not merely its inference cost:
\begin{equation}
MC(a) = C_{\text{compute}}(a) + \mathbb{E}[r(a) \cdot p(a)]
\label{eq:marginal_cost}
\end{equation}
where $r(a)$ denotes the liability risk (the cost of an error or safety violation) of an autonomous action $a$, and $p(a)$ denotes the probability of an incorrect execution. Even if $p(a)$ falls with continuous model improvement and $C_{\text{compute}}(a)$ approaches zero, $r(a)$ scales directly with the value of the underlying transaction. For high-value enterprise actions (contract execution, financial settlement, regulatory filings), the expected liability term $\mathbb{E}[r(a) \cdot p(a)]$ is orders of magnitude larger than the compute cost, creating a permanent floor on the zero-marginal-cost intuition that does not respond to Moore's Law. This formalizes the intuition in Agrawal et al. \cite{agrawal2018prediction} within the coordination cost framework developed in Section \ref{sec:model}. Furthermore, the true marginal cost of an agent is not compute, but \emph{liability}. As Agrawal et al. argued, while AI commoditizes prediction, it increases the value of judgment---specifically the liability for incorrect judgments \cite{agrawal2018prediction}. The risk of an agent executing a trade or signing a contract scales linearly with volume, creating a permanent floor on zero-marginal-cost scaling. Economically, this is an insurance premium. As agents move from ``drafting'' (low risk) to ``executing'' (high risk), the vendor must price in the cost of errors. Unlike compute, which follows Moore's Law, insurance premiums do not halve every 18 months; they scale with the value of the transaction.
\subsection{The Trust Boutique: Survival of the Middle}
\label{sec:trustboutique}

Finally, we predict a ``Missing Middle'' where mid-sized firms die. Critics argue that mid-sized firms survive on high-trust relationships that agents cannot replicate. We refine our prediction: the ``Missing Middle'' will not disappear, but transform into ``Trust Boutiques''---specialized liability wrappers. These firms will no longer sell execution (which is commoditized by agents) but accountability (the legal and moral liability for outcomes). They survive by becoming the ``moral crumple zone'' \cite{elish2019moral} for algorithmic systems, selling the human handshake that an agent cannot forge. However, effectively acting as an insurance wrapper requires far fewer humans than performing the execution itself. Thus, the firm size (headcount) still unbundles, even if the firm entity survives. (For a discussion on the wage implications of this shift, see the 'Scalable Boutique' analysis in Section 9.2).

\subsection*{Synthesis}

These four pathways are not fully independent. Platform capture at the waist (Section \ref{sec:thickwaist}) and liability concentration in trust boutiques (Section \ref{sec:trustboutique}) are particularly likely to co-occur: a thick-waist aggregator has strong incentives to also aggregate liability in order to extract rents from the execution layer, creating a self-reinforcing centralization dynamic. The remaining two pathways---cognitive load accumulation (Section \ref{sec:cognitiveload}) and the liability floor (Section \ref{sec:liability})---are more likely to operate independently and can be mitigated at the layer level without systemic restructuring: bounded malleability addresses the former; outcome-based pricing models address the latter. The durability of the hourglass equilibrium therefore depends most critically on preventing the co-occurrence of thick-waist capture and liability aggregation, since these two forces together would reconstitute a vertically integrated intermediary under a new name. If coordination remains compressed into standardized contracts and outcome verification, the hourglass persists as the stable architectural form of the agentic economy.

\section{Conclusion}

For most of economic history, large firms existed because coordination was expensive. Specialists needed managers; managers needed directors; directors needed strategy. Each layer added overhead but also reduced the cost of working across complex, interdependent tasks. AI agents challenge this logic at the root: if coordination can be compressed into a shared protocol and verification can scale with task volume rather than interaction complexity, the overhead that justified large integrated firms becomes optional.
The headless firm is the organizational form that follows from this change. Execution migrates outward to specialized agents; interaction coheres inward through a generative intent layer; in between sits a thin protocol waist that keeps the whole system composable. The resulting hourglass is not a design preference---it is an economic selection pressure. Firms that internalize broad capabilities in high-velocity domains face a ``complexity wall'' that compounds as domains accelerate. Firms that route execution through competitive agent markets face only the cost of verification and governance, which the model predicts can remain throughput-scaled.
The argument is conditional. If protocol layers thicken into proprietary gatekeepers, if verification becomes entangled with cross-agent liability, or if generative interfaces amplify rather than reduce cognitive load, the hourglass collapses and reintegration follows.
Ultimately, the Headless Firm is not a prediction about specific technologies or the trajectory of foundation models; it is a structural observation about economic selection pressure. As the underlying laws of coordination scaling change, the architectures that dominate the coming decade will be those that push execution to the specialized edges while compressing interaction into a generative, protocol-mediated waist.

\textbf{Limitations.} Our coordination cost model is stylized: the functional forms in equations (\ref{eq:coord_cost_decomp})--(\ref{eq:verification_coupling}) are chosen for tractability rather than empirical fit, and the exponent $\gamma$ is asserted to be less than one under the hourglass rather than derived from first principles. The Great Unbundling hypothesis (Hypothesis 1) is stated with domain-level scope conditions but has not been tested against longitudinal firm size data. The labor market predictions in Section \ref{sec:implications} rest on the scalable boutique mechanism, which depends on the assumption that high-leverage small firms can maintain bargaining power as agent execution commoditizes; this assumption may not hold if model providers extract value through inference pricing rather than capability differentiation. Finally, the analysis focuses on the DACH/EU enterprise context; regulatory differences (GDPR, AI Act compliance requirements) may shift the coordination cost landscape in ways not captured by our model.
The open research agenda is therefore not whether the Headless Firm is the right idea---it is whether the engineering and institutional conditions that stabilize it can be built fast enough to outpace the re-centralizing forces of platform capture, liability concentration, and governance complexity.

\bibliographystyle{plain}
\bibliography{references}

@article{acemoglu2022tasks,
 author = {Acemoglu, Daron and Restrepo, Pascual},
 doi = {10.3386/w28920},
 journal = {Econometrica},
 number = {5},
 pages = {1973--2016},
 title = {Tasks, automation, and the rise in US wage inequality},
 volume = {90},
 year = {2022}
}

@book{agrawal2018prediction,
 address = {Boston, MA},
 author = {Agrawal, Ajay and Gans, Joshua and Goldfarb, Avi},
 doi = {10.1080/15228053.2019.1673511},
 publisher = {Harvard Business Review Press},
 title = {Prediction Machines: The Simple Economics of Artificial Intelligence},
 year = {2018}
}

@misc{aiindex2025,
 author = {Stanford Institute for Human-Centered AI (HAI)},
 eprint = {2504.07139},
 publisher = {arXiv},
 title = {Artificial Intelligence Index Report 2025},
 year = {2025}
}

@inproceedings{akhshabi2011evolution,
 author = {Akhshabi, Saamer and Dovrolis, Constantinos},
 booktitle = {Proceedings of the ACM SIGCOMM 2011 Conference},
 doi = {10.1145/2018436.2018460},
 pages = {206--217},
 title = {The Evolution of Layered Protocol Stacks Leads to an Hourglass-Shaped Architecture},
 year = {2011}
}

@article{amodei2016concrete,
 author = {Amodei, Dario and Olah, Chris and Steinhardt, Jacob and Christiano, Paul and Schulman, John and Man{\'e}, Dan},
 journal = {arXiv preprint arXiv:1606.06565},
 title = {Concrete Problems in AI Safety},
 year = {2016}
}

@techreport{anthropic2024mcp,
 author = {Anthropic},
 doi = {10.59350/x8hvz-nxm60},
 institution = {Anthropic},
 note = {https://modelcontextprotocol.io},
 title = {Model Context Protocol: Specification and Architecture},
 year = {2024}
}

@inproceedings{apache2024iceberg,
 author = {Carlson, Ryan and others},
 booktitle = {Proceedings of the ACM SIGMOD International Conference on Management of Data},
 title = {Apache Iceberg: An Architectural Look Under the Covers},
 year = {2024}
}

@article{arxiv2025inference,
 author = {Zhuang, Boqin and Qiao, Jiacheng and Liu, Mingqian and Yu, Mingxing and Hong, Ping and Li, Rui},
 journal = {arXiv preprint arXiv:2510.26136},
 title = {Beyond Benchmarks: The Economics of {AI} Inference},
 year = {2025}
}

@article{autor2020superstar,
 author = {Autor, David and Dorn, David and Katz, Lawrence F. and Patterson, Christina and Van Reenen, John},
 doi = {10.1093/qje/qjaa004},
 journal = {The Quarterly Journal of Economics},
 number = {2},
 pages = {645--709},
 title = {The Fall of the Labor Share and the Rise of Superstar Firms},
 volume = {135},
 year = {2020}
}

@article{axtell2001zipf,
 author = {Axtell, Robert L.},
 doi = {10.1126/science.1062081},
 journal = {Science},
 number = {5536},
 pages = {1818--1820},
 title = {Zipf Distribution of {U}.{S}. Firm Sizes},
 volume = {293},
 year = {2001}
}

@book{baldwin2000design,
 author = {Baldwin, Carliss Y. and Clark, Kim B.},
 doi = {10.7551/mitpress/2366.001.0001},
 publisher = {MIT Press},
 title = {Design Rules: The Power of Modularity},
 year = {2000}
}

@article{barney1991firm,
 author = {Barney, Jay},
 doi = {10.1093/oso/9780199277681.003.0003},
 journal = {Journal of Management},
 number = {1},
 pages = {99--120},
 publisher = {Sage Publications},
 title = {Firm Resources and Sustained Competitive Advantage},
 volume = {17},
 year = {1991}
}

@book{brooks1975mythical,
 address = {Boston, MA},
 author = {Brooks, Frederick P.},
 doi = {10.1145/800027.808439},
 publisher = {Addison-Wesley},
 title = {The Mythical Man-Month: Essays on Software Engineering},
 year = {1975}
}

@article{brynjolfsson2000beyond,
 author = {Brynjolfsson, Erik and Hitt, Lorin M},
 doi = {10.1257/jep.14.4.23},
 journal = {Journal of Economic Perspectives},
 number = {4},
 pages = {23--48},
 title = {Beyond computation: Information technology, organizational transformation and business performance},
 volume = {14},
 year = {2000}
}

@techreport{brynjolfsson2023generative,
 author = {Brynjolfsson, Erik and Li, Danielle and Raymond, Lindsey R.},
 doi = {10.3386/w31161},
 institution = {National Bureau of Economic Research},
 number = {31161},
 title = {Generative AI at Work},
 type = {Working Paper},
 year = {2023}
}

@article{calvo1978supervision,
 author = {Calvo, Guillermo A. and Wellisz, Stanislaw},
 doi = {10.1086/260719},
 journal = {Journal of Political Economy},
 number = {5},
 pages = {943--952},
 title = {Supervision, Loss of Control, and the Optimum Size of the Firm},
 volume = {86},
 year = {1978}
}

@book{chappell2004enterprise,
 author = {Chappell, David},
 doi = {10.1007/springerreference_64100},
 publisher = {O'Reilly Media},
 title = {Enterprise Service Bus},
 year = {2004}
}

@misc{chen2017castles,
 author = {Chen, Jerry},
 howpublished = {Greylock Partners Blog},
 title = {Castles in the Cloud: The New Moats},
 url = {https://greylock.com/greymatter/castles-in-the-cloud/},
 year = {2017}
}

@misc{chen2025genui,
 archiveprefix = {arXiv},
 author = {Chen, Xiang ``Anthony'' and Knearem, Tiffany and Li, Yang},
 doi = {10.1145/3715336.3735780},
 eprint = {2501.13145},
 primaryclass = {cs.HC},
 title = {The {GenUI} Study: Exploring the Design of Generative {UI} Tools to Support {UX} Practitioners and Beyond},
 year = {2025}
}

@article{coase1937nature,
 author = {Coase, Ronald H.},
 doi = {10.1111/j.1468-0335.1937.tb00002.x},
 journal = {Economica},
 number = {16},
 pages = {386--405},
 title = {The Nature of the Firm},
 volume = {4},
 year = {1937}
}

@article{cobb2017growing,
 author = {Cobb, J. Adam and Lin, Ken-Hou},
 doi = {10.1287/orsc.2017.1125},
 journal = {Organization Science},
 number = {3},
 pages = {429--446},
 title = {Growing Apart: The Changing Firm-Size Wage Premium and Its Inequality Consequences},
 volume = {28},
 year = {2017}
}

@inproceedings{deng2023mind2web,
 author = {Deng, Xiang and others},
 booktitle = {NeurIPS},
 title = {Mind2Web: Towards a Generalist Agent for the Web},
 year = {2023}
}

@techreport{deutsch1994fallacies,
 author = {Deutsch, L Peter},
 doi = {10.1364/ofc.2020.th3k.2},
 institution = {Sun Microsystems},
 note = {Failures 1 through 7 were coined in 1994; the 8th was added later by James Gosling.},
 title = {The eight fallacies of distributed computing},
 year = {1994}
}

@article{eisenmann2006strategies,
 author = {Eisenmann, Thomas and Parker, Geoffrey and Van Alstyne, Marshall W.},
 journal = {Harvard Business Review},
 number = {10},
 pages = {92},
 title = {Strategies for Two-Sided Markets},
 volume = {84},
 year = {2006}
}

@article{elish2019moral,
 author = {Elish, Madeleine Clare},
 doi = {10.17351/ests2019.260},
 journal = {Engaging Science, Technology, and Society},
 pages = {40--60},
 publisher = {Society for the Social Studies of Science},
 title = {Moral Crumple Zones: Cautionary Tales in Human-Robot Interaction},
 volume = {5},
 year = {2019}
}

@book{evans2004domain,
 author = {Evans, Eric},
 doi = {10.5220/0005867905280535},
 publisher = {Addison-Wesley Professional},
 title = {Domain-Driven Design: Tackling Complexity in the Heart of Software},
 year = {2004}
}

@misc{casado2024unbundling,
  author       = {Martin Casado and Matt Bornstein},
  title        = {The Great Unbundling},
  howpublished = {Andreessen Horowitz Blog},
  year         = {2024},
  url          = {https://a16z.com/the-great-unbundling/}
}

@techreport{venkatesan2025services,
  author      = {Ashish Venkatesan and Nimit Jhunjhunwala},
  title       = {Stop Waiting for Proof: {Services-as-Software} Is Driving Non-Linear Growth},
  institution = {{HFS Research}},
  year        = {2025}
}

@misc{fersht2025saas,
  author       = {Phil Fersht},
  title        = {Service-as-Software: The New Paradigm},
  howpublished = {{HFS Research} Blog},
  year         = {2025},
  url          = {https://www.hfsresearch.com}
}

@article{fowler2014microservices,
 author = {Fowler, Martin and Lewis, James},
 journal = {MartinFowler.com},
 title = {Microservices: a definition of this new architectural style},
 url = {https://martinfowler.com/articles/microservices.html},
 year = {2014}
}

@article{gabaix2009power,
 author = {Gabaix, Xavier},
 doi = {10.3386/w14299},
 journal = {Annual Review of Economics},
 number = {1},
 pages = {255--294},
 title = {Power Laws in Economics and Finance},
 volume = {1},
 year = {2009}
}

@article{garicano2006organization,
 author = {Garicano, Luis and Rossi-Hansberg, Esteban},
 doi = {10.3386/w11458},
 journal = {The Quarterly Journal of Economics},
 number = {4},
 pages = {1383--1435},
 title = {Organization and inequality in a knowledge economy},
 volume = {121},
 year = {2006}
}

@techreport{gartner2020composable,
 author = {Kyte, Alexander and others},
 institution = {Gartner},
 title = {The Future of ERP is Composable},
 year = {2020}
}

@misc{gartner2025erp,
 author = {{Gartner}},
 doi = {10.1007/springerreference_6635},
 title = {Latest Enterprise Resource Planning (ERP) Insights},
 url = {https://www.gartner.com/en/information-technology/topics/enterprise-resource-planning},
 year = {2025}
}

@inproceedings{hadfield2019incomplete,
 author = {Hadfield-Menell, Dylan and Hadfield, Gillian K},
 booktitle = {Proceedings of the 2019 AAAI/ACM Conference on AI, Ethics, and Society},
 doi = {10.2139/ssrn.3165793},
 pages = {417--422},
 title = {Incomplete Contracting and AI Alignment},
 year = {2019}
}

@article{hart1990property,
 author = {Hart, Oliver and Moore, John},
 doi = {10.1086/261729},
 journal = {Journal of Political Economy},
 number = {6},
 pages = {1119--1158},
 title = {Property Rights and the Nature of the Firm},
 volume = {98},
 year = {1990}
}

@article{hassan2021daos,
 author = {Hassan, Samer and De Filippi, Primavera},
 doi = {10.14763/2021.2.1556},
 journal = {Internet Policy Review},
 number = {2},
 pages = {1--10},
 title = {Decentralized Autonomous Organization},
 volume = {10},
 year = {2021}
}

@article{hayek1945knowledge,
 author = {Hayek, Friedrich A.},
 doi = {10.2307/1809376},
 journal = {The American Economic Review},
 number = {4},
 pages = {519--530},
 title = {The Use of Knowledge in Society},
 volume = {35},
 year = {1945}
}

@article{henderson1990architectural,
 author = {Henderson, Rebecca M. and Clark, Kim B.},
 doi = {10.2307/2393549},
 journal = {Administrative Science Quarterly},
 number = {1},
 pages = {9--30},
 title = {Architectural Innovation: The Reconfiguration of Existing Product Technologies and the Failure of Established Firms},
 volume = {35},
 year = {1990}
}

@book{hohpe2003enterprise,
 author = {Hohpe, Gregor and Woolf, Bobby},
 doi = {10.1145/3561846.3561856},
 publisher = {Addison-Wesley Professional},
 title = {Enterprise Integration Patterns: Designing, Building, and Deploying Messaging Solutions},
 year = {2003}
}

@misc{idc2012darkdata,
 author = {Gantz, John and Reinsel, David},
 note = {IDC Analytical Study},
 publisher = {International Data Corporation},
 title = {The Digital Universe in 2020: Big Data, Bigger Digital Shadows, and Biggest Growth in the Far East},
 year = {2012}
}

@article{jensen1976theory,
 author = {Jensen, Michael C and Meckling, William H},
 doi = {10.1016/0304-405x(76)90026-x},
 journal = {Journal of Financial Economics},
 number = {4},
 pages = {305--360},
 title = {Theory of the firm: Managerial behavior, agency costs and ownership structure},
 volume = {3},
 year = {1976}
}

@book{jevons1865coal,
 author = {Jevons, William Stanley},
 doi = {10.2307/2339280},
 publisher = {Macmillan and Company},
 title = {The Coal Question: An Inquiry Concerning the Progress of the Nation, and the Probable Exhaustion of Our Coal-Mines},
 year = {1865}
}

@article{jones2009burden,
 author = {Jones, Benjamin F.},
 doi = {10.3386/w11360},
 journal = {The Review of Economic Studies},
 number = {1},
 pages = {283--317},
 publisher = {Oxford University Press},
 title = {The Burden of Knowledge and the "Death of the Renaissance Man": Is Innovation Getting Harder?},
 volume = {76},
 year = {2009}
}

@article{karpathy2017software2,
 author = {Karpathy, Andrej},
 journal = {Medium},
 title = {Software 2.0},
 year = {2017}
}

@article{langlois2002modularity,
 author = {Langlois, Richard N},
 doi = {10.4337/9781843767107.00009},
 journal = {Journal of Economic Behavior \& Organization},
 number = {1},
 pages = {19--37},
 title = {Modularity in technology and organization},
 volume = {49},
 year = {2002}
}

@techreport{leviathan2025generative,
 author = {Leviathan, Yaniv and others},
 institution = {Google Research},
 title = {Generative UI: LLMs are effective UI generators},
 url = {https://generativeui.github.io/},
 year = {2025}
}

@inproceedings{liu2024agentbench,
 author = {Liu, Xiao and others},
 booktitle = {ICLR},
 title = {AgentBench: Evaluating LLMs as Agents},
 year = {2024}
}

@article{lucas1978size,
 author = {Lucas, Robert E.},
 doi = {10.2307/3003596},
 journal = {The Bell Journal of Economics},
 number = {2},
 pages = {508--523},
 title = {On the Size Distribution of Business Firms},
 volume = {9},
 year = {1978}
}

@article{malone1994interdisciplinary,
 author = {Malone, Thomas W. and Crowston, Kevin},
 doi = {10.4324/9781410605863-6},
 journal = {ACM Computing Surveys},
 number = {1},
 pages = {87--119},
 title = {The Interdisciplinary Study of Coordination},
 volume = {26},
 year = {1994}
}

@techreport{martens2020digital,
 author = {Martens, Bertin and others},
 doi = {10.2139/ssrn.3164170},
 institution = {JRC Digital Economy Working Paper 2018-02},
 title = {The digital transformation of news media and the rise of disinformation and fake news},
 year = {2020}
}

@inproceedings{mei2025aios,
 author = {Mei, Kai and others},
 booktitle = {Conference on Language Modeling (COLM)},
 title = {AIOS: LLM Agent Operating System},
 year = {2025}
}

@article{moore2011systems,
 author = {Moore, Geoffrey},
 journal = {AIIM White Paper},
 publisher = {AIIM},
 title = {Systems of engagement and the future of enterprise {IT}},
 year = {2011}
}

@article{murray2021algorithms,
 author = {Murray, Alex and Rhymer, Jen and Sirmon, David G.},
 doi = {10.1016/j.hrmr.2021.100838},
 journal = {Human Resource Management Review},
 number = {2},
 title = {Algorithms as Work Designers: How Algorithmic Management Influences the Design of Jobs},
 volume = {31},
 year = {2021}
}

@article{openapps2026,
 author = {Ullrich, Karen and Su, Jingtong and others},
 journal = {arXiv preprint arXiv:2511.20766},
 title = {OpenApps: Simulating Environment Variations to Measure UI Agent Reliability},
 year = {2025}
}

@book{parker2016platform,
 author = {Parker, Geoffrey G. and Van Alstyne, Marshall W. and Choudary, Sangeet Paul},
 publisher = {W. W. Norton \& Company},
 title = {Platform Revolution: How Networked Markets Are Transforming the Economy and How to Make Them Work for You},
 year = {2016}
}

@techreport{postel1981tcp,
 author = {Postel, Jon},
 doi = {10.17487/rfc0793},
 institution = {RFC 793, Internet Engineering Task Force},
 month = {September},
 note = {The Robustness Principle is formulated in Section 2.10},
 title = {Transmission Control Protocol},
 year = {1981}
}

@article{puranam2014new,
 author = {Puranam, Phanish and Alexy, Oliver and Reitzig, Markus},
 doi = {10.1093/oso/9780199672363.003.0008},
 journal = {Academy of Management Review},
 number = {2},
 pages = {162--180},
 title = {What's ``New'' about New Forms of Organizing?},
 volume = {39},
 year = {2014}
}

@article{qin2023toolllm,
 author = {Yujia Qin and Shihao Liang and Yining Ye and Kunlun Zhu and Lan Yan and Yaxi Lu and Yankai Lin and Xin Cong and Xiangru Tang and Bill Qian and Sihan Zhao and Lauren Hong and Runchu Tian and Ruobing Xie and Jie Zhou and Mark Gerstein and Dahai Li and Zhiyuan Liu and Maosong Sun},
 journal = {arXiv preprint arXiv:2307.16789},
 title = {ToolLLM: Facilitating Large Language Models to Master 16000+ Real-world APIs},
 year = {2023}
}

@book{rifkin2014zero,
 author = {Rifkin, Jeremy},
 doi = {10.5325/utopianstudies.26.2.0422},
 publisher = {Palgrave Macmillan},
 title = {The Zero Marginal Cost Society: The Internet of Things, the Collaborative Commons, and the Eclipse of Capitalism},
 year = {2014}
}

@article{rochet2003platform,
 author = {Rochet, Jean-Charles and Tirole, Jean},
 doi = {10.1162/154247603322493212},
 journal = {Journal of the European Economic Association},
 number = {4},
 pages = {990--1029},
 title = {Platform competition in two-sided markets},
 volume = {1},
 year = {2003}
}

@article{roth2002economist,
 author = {Roth, Alvin E.},
 doi = {10.1111/1468-0262.00335},
 journal = {Econometrica},
 number = {4},
 pages = {1341--1378},
 title = {The Economist as Engineer: Game Theory, Experimentation, and Computation as Tools for Design Economics},
 volume = {70},
 year = {2002}
}

@article{saltzer1984end,
 author = {Saltzer, Jerome H. and Reed, David P. and Clark, David D.},
 doi = {10.1145/357401.357402},
 journal = {ACM Transactions on Computer Systems (TOCS)},
 number = {4},
 pages = {277--288},
 publisher = {ACM},
 title = {End-to-end arguments in system design},
 volume = {2},
 year = {1984}
}

@article{schick2024toolformer,
 author = {Schick, Timo and others},
 journal = {Advances in Neural Information Processing Systems},
 title = {Toolformer: Language Models Can Teach Themselves to Use Tools},
 year = {2024}
}

@techreport{shahidi2025coasean,
 author = {Shahidi, Peyman and Rusak, Gili and Manning, Benjamin S. and Fradkin, Andrey and Horton, John J.},
 doi = {10.3386/w34468},
 institution = {National Bureau of Economic Research},
 note = {Presented at the 2025 Economics of Transformative AI Workshop},
 number = {34468},
 title = {The Coasean Singularity? Demand, Supply, and Market Design with {AI} Agents},
 type = {Working Paper},
 year = {2025}
}

@book{shapiro1998information,
 address = {Boston, MA},
 author = {Shapiro, Carl and Varian, Hal R.},
 doi = {10.1086/603169},
 publisher = {Harvard Business Press},
 title = {Information Rules: A Strategic Guide to the Network Economy},
 year = {1998}
}

@misc{shazeer2017moe,
 archiveprefix = {arXiv},
 author = {Noam Shazeer and Azalia Mirhoseini and Krzysztof Maziarz and Andy Davis and Quoc Le and Geoffrey Hinton and Jeff Dean},
 doi = {10.1109/ijcnn54540.2023.10191904},
 eprint = {1701.06538},
 primaryclass = {cs.LG},
 title = {Outrageously Large Neural Networks: The Sparsely-Gated Mixture-of-Experts Layer},
 year = {2017}
}

@article{simon1955behavioral,
 author = {Simon, Herbert A},
 doi = {10.7249/p365},
 journal = {The Quarterly Journal of Economics},
 number = {1},
 pages = {99--118},
 title = {A behavioral model of rational choice},
 volume = {69},
 year = {1955}
}

@book{skelton2019team,
 author = {Skelton, Matthew and Pais, Manuel},
 publisher = {IT Revolution},
 title = {Team Topologies: Organizing Business and Technology Teams for Fast Flow},
 year = {2019}
}

@article{snell2024scaling,
 author = {Snell, Charlie and Lee, Jaehoon and Xu, Kelvin and Kumar, Aviral},
 journal = {arXiv preprint arXiv:2408.03314},
 title = {Scaling LLM Test-Time Compute Optimally can be more Effective than Scaling Model Parameters},
 year = {2024}
}

@article{song2019firming,
 author = {Song, Jae and Price, David J. and Guvenen, Fatih and Bloom, Nicholas and von Wachter, Till},
 doi = {10.3386/w21199},
 journal = {The Quarterly Journal of Economics},
 number = {1},
 pages = {1--50},
 title = {Firming Up Inequality},
 volume = {134},
 year = {2019}
}

@article{stigler1951division,
 author = {Stigler, George J.},
 doi = {10.1086/257075},
 journal = {Journal of Political Economy},
 number = {3},
 pages = {185--193},
 publisher = {The University of Chicago Press},
 title = {The Division of Labor is Limited by the Extent of the Market},
 volume = {59},
 year = {1951}
}

@book{sundararajan2016sharing,
 address = {Cambridge, MA},
 author = {Sundararajan, Arun},
 doi = {10.1177/0972262917712390},
 publisher = {MIT Press},
 title = {The Sharing Economy: The End of Employment and the Rise of Crowd-Based Capitalism},
 year = {2016}
}

@article{sweller1988cognitive,
 author = {Sweller, John},
 doi = {10.1207/s15516709cog1202_4},
 journal = {Cognitive Science},
 number = {2},
 pages = {257--285},
 title = {Cognitive Load During Problem Solving: Effects on Learning},
 volume = {12},
 year = {1988}
}

@book{tainter1988collapse,
 author = {Tainter, Joseph A.},
 doi = {10.2307/505333},
 publisher = {Cambridge University Press},
 title = {The Collapse of Complex Societies},
 year = {1988}
}

@incollection{thomas2002disposable,
 author = {Thomas, Dave},
 booktitle = {Beautiful Architecture: Leading Thinkers Reveal the Hidden Beauty in Software Design},
 editor = {Spinellis, Diomidis and Gousios, Georgios},
 publisher = {O'Reilly Media},
 title = {The Case For Disposable Software: `Big Architecture' Not Always Essential},
 year = {2009}
}

@article{tomasev2026intelligent,
 author = {Toma{\v{s}}ev, Nenad and Franklin, Matija and Osindero, Simon},
 journal = {arXiv preprint arXiv:2602.11865},
 title = {Intelligent AI Delegation},
 url = {https://arxiv.org/abs/2602.11865},
 year = {2026}
}

@article{trask2023data,
 author = {Trask, Andrew and others},
 journal = {Advances in Neural Information Processing Systems},
 title = {Data Moats and Proprietary Training Sets in Foundation Models},
 year = {2023}
}

@article{vanvalen1973redqueen,
 author = {Van Valen, Leigh},
 doi = {10.1007/s13752-021-00391-w},
 journal = {Evolutionary theory},
 pages = {1--30},
 title = {A new evolutionary law},
 volume = {1},
 year = {1973}
}

@techreport{w3c2022did,
 author = {Sporny, Manu and Longley, Dave and Sabadello, Markus and Reed, Drummond and Steele, Orie and Allen, Christopher},
 doi = {10.1109/icufn55119.2022.9829562},
 institution = {W3C},
 title = {Decentralized Identifiers (DIDs) v1.0},
 type = {Recommendation},
 url = {https://www.w3.org/TR/did-core/},
 year = {2022}
}

@inproceedings{wang2024mome,
 author = {Shen, Leyang and Chen, Gongwei and Shao, Rui and Guan, Weili and Nie, Liqiang},
 booktitle = {Advances in Neural Information Processing Systems (NeurIPS)},
 doi = {10.52202/079017-1330},
 title = {MoME: Mixture of Multimodal Experts for Generalist Multimodal Large Language Models},
 year = {2024}
}

@techreport{wang2025agentic,
 author = {Wang, Yu and Lu, Y.},
 institution = {Microsoft Research},
 number = {MSR-TR-2025-62},
 title = {Agentic Media: Reimagining the future of communication},
 year = {2025}
}

@misc{west2017scale,
 author = {West, Geoffrey},
 doi = {10.1177/1473095220914159},
 publisher = {Penguin},
 title = {Scale: The Universal Laws of Growth, Innovation, Sustainability, and the Pace of Life in Organisms, Cities, Economies, and Companies},
 year = {2017}
}

@article{williamson1967hierarchical,
 author = {Williamson, Oliver E.},
 doi = {10.1086/259258},
 journal = {Journal of Political Economy},
 number = {2},
 pages = {123--138},
 title = {Hierarchical Control and Optimum Firm Size},
 volume = {75},
 year = {1967}
}

@article{williamson1981economics,
 author = {Williamson, Oliver E},
 doi = {10.1086/227496},
 journal = {American Journal of Sociology},
 number = {3},
 pages = {548--577},
 title = {The economics of organization: The transaction cost approach},
 volume = {87},
 year = {1981}
}

@article{wuchty2007increasing,
 author = {Wuchty, Stefan and Jones, Benjamin F and Uzzi, Brian},
 doi = {10.1126/science.1136099},
 journal = {Science},
 number = {5827},
 pages = {1036--1039},
 title = {The increasing dominance of teams in production of knowledge},
 volume = {316},
 year = {2007}
}

@inproceedings{xia2024generative,
 author = {Xia, Haijun},
 booktitle = {Proceedings of the 2024 CHI Conference on Human Factors in Computing Systems (Workshop on GenAI for UI)},
 doi = {10.1145/3706598.3713285},
 title = {Generative and Malleable User Interfaces},
 year = {2024}
}

@inproceedings{zhou2024webarena,
 author = {Zhou, Shuyan and others},
 booktitle = {ICLR},
 title = {WebArena: A Realistic Web Environment for Building Autonomous Agents},
 year = {2024}
}

\end{document}